\documentclass[10pt,twocolumn,letterpaper]{article}

\usepackage[T1]{fontenc}         
\usepackage[pagenumbers]{cvpr}

\usepackage{float}
\usepackage{nicefrac}
\usepackage{microtype}

\definecolor{cvprblue}{rgb}{0.21,0.49,0.74}
\usepackage[pagebackref,breaklinks,colorlinks,allcolors=cvprblue]{hyperref}

\def\confName{CVPR}
\def\confYear{2027}

\title{Sub-Meter Canopy Height Synthesis from Sentinel-2 via Anchored Residual Flow}
\author{Kiarie Ndegwa$^{1}$ \quad Tony Chang$^{1}$ \quad Andreas Gros$^{2}$ \quad Vincent A. Landau$^{1}$\\
David D. Diaz$^{1}$ \quad Luke Zachmann$^{1}$ \quad Nathan E. Rutenbeck$^{1}$ \quad Guy C Bayes$^{3}$\\[0.35em]
$^{1}$Vibrant Planet \quad $^{2}$Ar\^{e}te Glaciers Initiative \quad $^{3}$Terradot\\
}

\begin{document}
 \maketitle

\begin{abstract}
Canopy height models (CHMs) - per-pixel maps of vegetation top height - are essential for forest carbon accounting, wildfire fuel mapping, and tree-level ecology, yet sub-meter CHMs require airborne LiDAR refreshed on multi-year national cycles. We produce 0.5 m CHMs from free 10 m Sentinel-2 imagery, a 20\(\times\) cross-modal super-resolution task where optical reflectance must predict geometric height it does not directly observe. Motivated by the perception-distortion trade-off, we derive two operating points with shared conditioning. A deterministic regressor (Anchor) targets low distortion; a conditional flow matching model, trained on the Anchor's residual in a learned CHM latent space (AR-flow), targets perceptual fidelity. Trained on 197k LiDAR-paired tiles spanning 69 ecoregions across the contiguous United States and evaluated on 2,530 held-out tiles against eight baselines, the Anchor achieves 2.83 m MAE using only free satellite input. The AR-flow reduces sliced Wasserstein distance from the Anchor's 1.13 to 0.208, at a higher MAE of 3.68 m. Crown-scale statistics improve, but strict individual-tree detection remains weak: generated sub-pixel detail represents plausible structure rather than measured individual trees. Quarterly continental refresh becomes tractable, letting satellite monitoring inform timing and extent for more expensive data acquisitions.
\end{abstract}

\section{Introduction}
\label{sec:intro}

\begin{figure}[t]
\centering
\includegraphics[width=\linewidth]{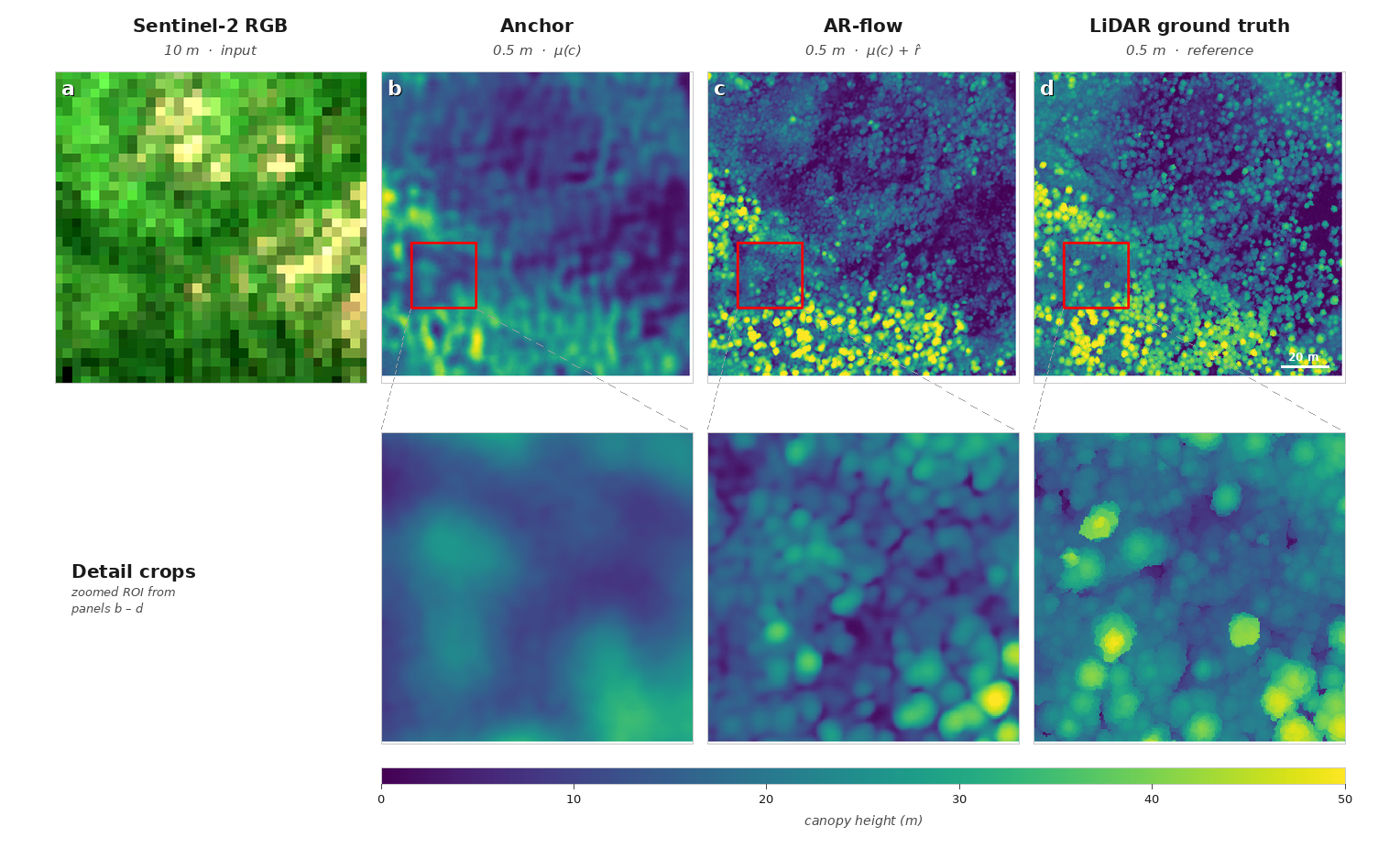}
\caption{\textbf{Sub-meter canopy height from 10~m Sentinel-2.} Top: S2 RGB (10~m). Middle: model outputs and LiDAR at 0.5~m. Bottom: crown-level crops. The Anchor produces smooth surfaces; the AR-flow  synthesizes crown-scale structure. Klamath Mountains conifer tile, OR. Viridis 0--50~m.}
\label{fig:hero}
\end{figure}

\subsection{Application and Challenge}

Canopy Height Models (CHMs) -- per-pixel vegetation-top height -- are essential for carbon accounting, wildfire fuel mapping, and tree-level ecology. Sub-meter CHMs require airborne LiDAR at \textasciitilde\$100--300/km$^2$; USGS 3DEP set out to complete national 3D elevation collection in eight years \citep{sugarbaker2014}. Satellite-derived CHM products include Meta CHMv2 \citep{brandt2026meta}, Planet Forest Carbon Monitoring \citep{planet_forest_carbon_monitoring}, and CTrees \citep{ctrees_tree_level}. Meta CHMv2, the strongest baseline in our evaluation, uses commercial Maxar imagery. Sentinel-2 acquires globally every 5 days at 10~m, at zero marginal cost.

Mapping S2 imagery to a sub-meter CHM requires predicting geometric height from optical reflectance. The target is a scalar height field, but accurate average heights and plausible crown structure impose different demands on the prediction. This is an instance of the distortion-perception trade-off \citep{blau2018perception}: carbon inventory requires calibrated mean height, whereas crown delineation requires fine-scale structure. We provide two operating points and outperform the commercial-input baseline on the reported perception metrics using free input.

\subsection{Approach and Roadmap}

We decompose the problem into two stages sharing Sentinel-2 and frozen DINOv3 conditioning. \textbf{Stage~1} trains a deterministic regressor in a pretrained CHM autoencoder latent space. This regressor, the Anchor, targets the minimum mean squared error (MMSE) estimate. \textbf{Stage~2} learns a flow on the residual between the frozen Anchor and the ground-truth latent. This anchored residual flow (AR-flow) supplies the structural variation lost through regression. The two outputs emphasize height accuracy and perceptual fidelity, respectively. Section~\ref{sec:method} describes both stages. We evaluate encoder and flow choices at matched compute (Sections~\ref{sec:encoder_ablation}--\ref{sec:flow_ablation}), compare ten products (Sections~\ref{sec:eco_analysis}--\ref{sec:sota}), and assess crown structure and seasonal cadence (Sections~\ref{sec:crown}--\ref{sec:seasonal}).

\subsection{Contributions}

\begin{enumerate}
\item \textbf{Continental-scale sub-meter CHM from free S2.} Trained on 197k LiDAR tiles spanning 69/72 CONUS ecoregions, evaluated against eight baselines with per-ecoregion paired bootstrap.

\item \textbf{Explicit limits on sub-pixel inference.} Reflectance provides context for height prediction at the 10~m input scale, but does not directly measure geometric height. Within a pixel footprint, different tree arrangements can produce the same spectral signature. Sub-10~m structure is therefore context-conditioned synthesis rather than measurement. ``Sub-meter'' throughout refers to output sampling, never to error magnitude.

\item \textbf{Anchored residual flow yielding two operating points.} The Anchor achieves comparable cohort-mean MAE to the strongest commercial-input baseline; the AR-flow improves perceptual metrics, including LPIPS, which is not its training objective. Matched-compute ablations isolate encoder, transport, and anchor factors.

\item \textbf{Finding: sub-meter sampling does not yield sub-meter structure.} In a 16-tile spectral probe at identical 0.5~m output the Anchor retains 0.008 of LiDAR's structural power at the 10~m input scale against the AR-flow's 0.95, and is furthest from LiDAR on crown height, size and spatial arrangement. The AR-flow's structure is synthesised, not recovered: it under-clusters relative to LiDAR and its injected detail lowers Block R$^2$.
\end{enumerate}

\textbf{Scope.} This is an empirical study of canopy-height prediction; model weights remain institutional IP and are not released. Appendices cited throughout the paper are provided in the supplementary material.

\section{Related Work}
\label{sec:related}

\textbf{Latent generative SR.} LDM \citep{rombach2022high} established super-resolution in VAE latent space. Related restoration methods include ResShift's residual chains \citep{yue2023resshift}, I2SB's Schr\"{o}dinger bridges \citep{liu2023i2sb}, and LBM's bridge matching \citep{chadebec2025lbm}. These methods operate on natural-image autoencoder latents, or on pixels for I2SB, with inputs and outputs in the same modality. We use a domain-specific CHM autoencoder with cross-modal conditioning.

\textbf{Two-stage deterministic + generative pipelines.} The perception-distortion trade-off prevents a single estimator from simultaneously minimizing distortion and maximizing perception \citep{blau2018perception}. \citet{freirich2021theory} showed that minimum MSE at perfect perceptual quality can be obtained by optimally transporting the MMSE estimate. PMRF \citep{ohayon2025pmrf} approximates this transport with rectified flow, and Latent-PMRF \citep{luo2025latentpmrf} performs it in a learned latent space. Both restore degraded images within the same modality; our Anchor predicts height from reflectance.

CorrDiff \citep{mardani2024corrdiff} and ELIR \citep{cohen2025elir} also combine deterministic prediction with generative refinement. RDDM \citep{liu2024rddm} and InDI \citep{delbracio2023inversion} use related residual or iterative restoration methods without a separately trained deterministic first stage. AR-flow combines a frozen first stage with conditional residual generation in a CHM latent space.

\textbf{Cross-modal remote-sensing SR.} \citet{kim2025cbbdm} and C-DiffSET \citep{do2026cdiffset} use latent diffusion for SAR-to-optical; DiffusionSat \citep{khanna2024diffusionsat} is a satellite foundation model but not two-stage. Concurrently, \citet{hester2026flow} apply single-stage flow matching to 4$\times$ Sentinel-2 super-resolution over CONUS and trade perception against distortion at inference, but super-resolve reflectance to imagery rather than predicting height. These methods do not address scalar canopy height at 20x cross-modal upsampling.

Regression products (Meta CHMv2 \citep{brandt2026meta}, ETH \citep{lang2023high}, GLAD \citep{potapov2021glad}, \citealt{pauls2024canopy}) map satellite imagery to canopy height without a generative component. Two concurrent works do super-resolve beyond Sentinel-2's native grid: SERA-H \citep{boudras2025serah} couples an EDSR super-resolution module to a UTAE temporal encoder over S1+S2 time series, and THREASURE-Net \citep{kalinicheva2025threasure} regresses and super-resolves jointly from S2 time series against LiDAR HD, both reaching 2.5~m over France at \textasciitilde2.6~m MAE. Both use regression and produce 2.5~m maps; we sample at 0.5~m across CONUS and add a generative operating point, with the pixel-accuracy cost quantified in Appendix~\ref{sec:block_r2}.

The earlier VibrantSR flow-matching model \citep{ndegwa2026vibrantsr} covers 22 western ecoregions at 4.39~m MAE on a different cohort and height mask; the present work adds the deterministic anchor, the residual formulation, and 69-ecoregion CONUS coverage. Section~\ref{sec:flow_ablation} compares direct and residual targets using the present architecture, not the published VibrantSR checkpoint.

\textbf{Positioning.} AR-flow combines cross-modal conditioning, a frozen MSE first stage, and latent-space residual flow. Appendix~\ref{sec:positioning_appendix} compares the closest methods along these three axes.

\section{Method}
\label{sec:method}

The model takes a 12-band S2 composite (48$\times$48 px, 480$\times$480~m) and outputs a 0.5~m CHM over the same 480~m footprint (960$\times$960 px). The shared conditioning supports a deterministic prediction from the Anchor and a generative prediction from AR-flow. The conditioning bundle is $c = \{S2_{12}, \mathrm{DINOv3}\}$ with DINOv3 \citep{simeoni2025dinov3} as the frozen ViT-L/16 SAT-493M. A frozen DC-AE \citep{chen2024dcae} maps the CHM target to $z_y \in \mathbb{R}^{128 \times 15 \times 15}$ (64x downsampling). Stage~1 predicts $\mu_\theta(c)$, and Stage~2 learns a flow on the residual $r = z_y - \mu_\theta(c)$. At inference, the sampled residual is added to the Anchor and decoded through the frozen CHM autoencoder: $\hat z_y = \mu_\theta(c) + \hat r$. The decomposition is motivated by \citet{freirich2021theory}: minimum MSE at perfect perceptual quality can be obtained by optimally transporting the MMSE estimate.

\begin{figure*}[t]
\centering
\includegraphics[width=\textwidth]{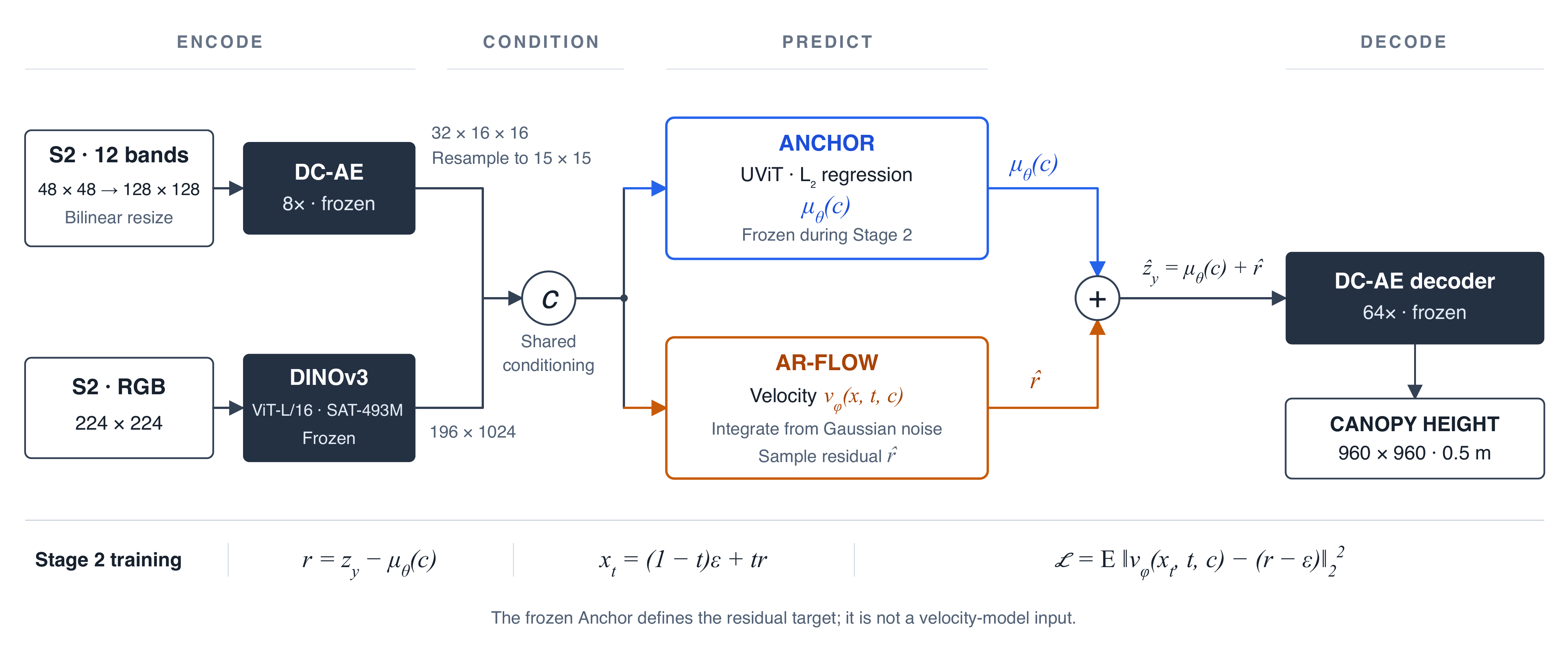}
\caption{\textbf{System diagram.}  Frozen encoders supply the shared conditioning bundle $c = \{S2_{12}, \mathrm{DINOv3}\}$.  The Stage-1 UViT L2 regressor  produces anchor $\mu(c)$.  Stage-2 AR-flow  uses $x_t = (1{-}t)\varepsilon + t\,r$, $r = z_y - \mu(c)$. At inference $\hat z_y = \mu(c) + \hat r$ is decoded to 960$\times$960 CHM.}
\label{fig:system}
\end{figure*}

\subsection{Conditioning Encoders}

The conditioning bundle $c$ has two streams: (1) the 12-band S2 composite, bilinearly upsampled from $48 \times 48$ to $128 \times 128$ and encoded by a frozen DC-AE (8x downsampling, 32 channels) into a $32 \times 16 \times 16$ latent resampled to the CHM grid; (2) frozen DINOv3 ViT-L/16 SAT-493M patch tokens (196 tokens, dim 1024) entering via cross-attention at three UViT decoder depths. DINOv3's patch embedding is three-channel, so this stream sees only the RGB bands (B04/B03/B02) resized to $224 \times 224$; the red-edge, NIR and SWIR bands reach the model through the S2 autoencoder stream alone. DINOv3 matches the encoder used by Meta CHMv2 \citep{brandt2026meta}; two alternatives (Clay v1.5, AlphaEarth) are ablated in Section~\ref{sec:encoder_ablation}. 

Both autoencoders follow DC-AE \citep{chen2024dcae}. They are trained from scratch on domain-specific data with L1, LPIPS and adversarial losses, then frozen. They have different compression factors: the S2 side-channel encoder is $f{=}8$, whereas the generative bottleneck is the CHM autoencoder at $f{=}64$, which reduces a $960 \times 960$ tile to $15 \times 15$ spatial positions (225 tokens, $4{,}096\times$ fewer). Both autoencoders and the UViT are fully specified in Appendix~\ref{sec:architecture_appendix}.

\subsection{Stage 1: Deterministic Cross-Modal Regressor (Anchor)}

Stage~1 is a UViT regressing from $c$ to $z_y$ in the CHM latent ($128 \times 15 \times 15$), with pure L2 loss:
\[
\mathcal{L}_{\text{stage 1}}(\theta) \;=\; \mathbb{E}_{(c, y) \sim \mathcal{D}} \;\big\| \mu_\theta(c) - z_y \big\|_2^2 .
\]
The population L2 minimizer is the conditional mean $\mathbb{E}[z_y \mid c]$, which is optimal for latent squared error. Finite training need not attain this optimum, and latent optimality does not guarantee optimal decoded MAE or Block R$^2$. Because the CHM autoencoder was trained with perceptual and adversarial losses, even this conditional-mean latent decodes to a CHM that preserves gross canopy structure -- though it smooths fine-grained crown detail, motivating Stage~2. Once trained, Stage~1 is frozen and serves as both a standalone regression product (the Anchor) and the reference point for the AR-flow residual.

\subsection{Stage 2: Anchor-Residual Flow (AR-flow)}

Stage~2 uses conditional flow matching \citep{liu2023flow,lipman2023flow} on the residual $r = z_y - \mu_\theta(c)$. With $\varepsilon \sim \mathcal{N}(0, I)$ and $t \in [0,1]$, the linear-path interpolant is $x_t = (1-t)\varepsilon + t\,r$ and the velocity head $v_\phi(x_t, t, c)$ is trained by
\[
\mathcal{L}_{\text{AR-flow}}(\phi) = \mathbb{E}\,\big\| v_\phi(x_t, t, c) - (r - \varepsilon) \big\|_2^2 .
\]

At inference we integrate $dx/dt = v_\phi(x, t, c)$ from $\varepsilon$ to $\hat r$ and reconstruct $\hat z_y = \mu_\theta(c) + \hat r$, decoded through the frozen autoencoder. The anchor $\mu_\theta(c)$ enters only through the residual target $r$ during training, never as a concatenated input to the velocity head.

\subsection{Design Choices}

\textbf{(1) Latent-space regression.} Both stages operate in the CHM latent ($15{\times}15$, 128~ch) rather than the $960{\times}960$ pixel grid, reducing the number of spatial positions by \textasciitilde4100x (\textasciitilde3~s/tile on A10G). \textbf{(2) Frozen Stage~1.} Pure L2 training targets the latent conditional mean and yields a standalone product. \textbf{(3) Residual target without anchor concatenation.} The velocity head receives $(x_t, t, c)$ only; the anchor influences Stage~2 solely through the training target $r = z_y - \mu_\theta(c)$.

\section{Experimental Setup}
\label{sec:setup}

\subsection{Dataset}

\textbf{CHM targets.} 480$\times$480~m tiles at \textasciitilde0.5~m GSD (960$\times$960 px), derived from USGS 3DEP airborne LiDAR and stratified by EPA Level-III ecoregion \citep{omernik1987ecoregions}. The main training set (\textasciitilde197k tiles) spans 69/72 forested ecoregions; ablations use the smaller year-aligned subset (Section~\ref{sec:temporal}). Deterministic hashing on tile coordinates makes tile IDs disjoint; nearby tiles may remain spatially correlated.

\textbf{Sentinel-2 inputs.} Cloud-masked three-month median composite of S2 L2A surface reflectance (12 bands, 10~m GSD). The compositing window is the season closest to the LiDAR acquisition date with strict year alignment; the seasonal ablation (Section~\ref{sec:seasonal}) evaluates all four quarters.

\textbf{DINOv3 features.} The frozen DINOv3 ViT-L/16 SAT-493M \citep{simeoni2025dinov3} produces 196 patch tokens from the RGB bands of the S2 composite, resized to $224 \times 224$. DINOv3 matches the checkpoint used by Meta CHMv2; inputs, training data, and downstream recipes still differ in Section~\ref{sec:sota}.

\begin{figure*}[t]
\centering
\includegraphics[width=0.75\textwidth]{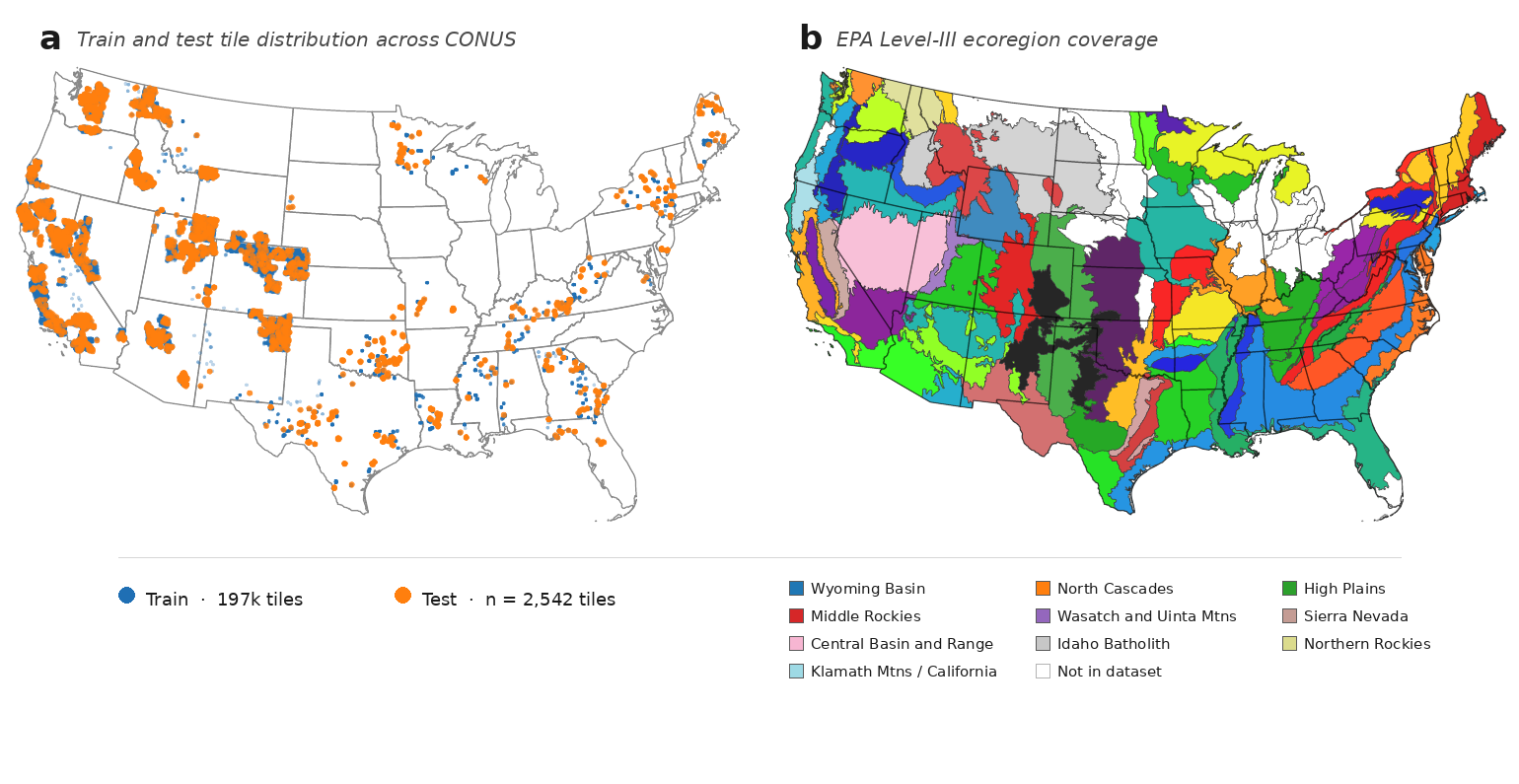}
\caption{\textbf{CONUS tile coverage and ecoregion context.} (a) Train ($n{=}197{,}257$, blue) and test ($n{=}2{,}542$, orange) tiles. (b) EPA Level~III ecoregions; 72 covered by our dataset are colored; 13 outside rendered white. Test cohort spans 69 ecoregions (96\% coverage).}
\label{fig:coverage}
\end{figure*}

\textbf{Temporal alignment.}
\label{sec:temporal}
To prevent temporal confounding from forest change between S2 acquisition and LiDAR collection, our primary evaluation protocol enforces strict year alignment (\texttt{s2\_acq\_year == chm\_acq\_year}), yielding 4,299 training and 2,542 test tiles (2,530 after excluding tiles missing baseline products). All headline numbers in Section~\ref{sec:results} report this year-aligned cohort; the main model additionally trains on the full \textasciitilde197k-tile corpus without year constraints. The evaluation prioritizes reference density (\textasciitilde3.6M LiDAR pixels per km$^2$ across 69 ecoregions) over geographic breadth, complementing global GEDI-based evaluations (\textasciitilde25 footprints per km$^2$).

\subsection{Architecture}
\label{sec:architecture}

Component-by-component shapes, parameter counts and recipes are given in Appendix~\ref{sec:architecture_appendix}. Both stages use the same UViT-style backbone \citep{bao2023uvit} (embed 768, depth 12, 12 heads) with added DINOv3 cross-attention at three decoder depths. All frozen components remain fixed throughout.

\subsection{Training}
\label{sec:training}

We use AdamW with learning rate $1 \times 10^{-4}$, weight decay 0.03, cosine schedule with 50\% warmup and minimum learning rate $10^{-7}$, and gradient clipping at 1.0. Global batch size is 16 in full precision. Each variant is trained from scratch; Stage~1 is frozen before Stage~2 begins. Each ablation run takes approximately one day on a single A10G (\textasciitilde15 runs total). The main AR-flow trained for 98 epochs on the \textasciitilde197k-tile pan-CONUS corpus on 8xA100 over approximately four days. 

Inference uses a Dormand-Prince adaptive-step solver \citep[dopri5;][]{dormand1980dopri}; forward Euler ($N{=}32$) is used for training-time validation. Each reported AR-flow result uses a single draw per tile with a fixed sampling seed. This makes the reported draws reproducible; the prediction rasters intended for release are the same draws used to compute the metrics. Multi-sample behaviour is examined separately in Appendix~\ref{sec:uq_appendix}.

\subsection{Evaluation Metrics}

The perception-distortion trade-off means no single metric captures both operating points. We report distortion and perception separately, using three perception metric families: optimal transport, wavelets, and learned features. Agreement across them helps assess whether a result depends on a single metric. Height errors are computed in meters against the ground-truth LiDAR CHM on the held-out test set; perceptual metrics use their own scales.

Distortion: \textbf{MAE} and \textbf{RMSE} (meters); \textbf{Edge error (EE)}, the mean absolute difference between Sobel gradient magnitudes of prediction and ground truth, normalized per tile (adapted from \citealt{tolan2024canopy}). Perception: \textbf{SWD}, sliced Wasserstein distance on $128{\times}128$ patches (adapted from \citealt{karras2018progressive}); \textbf{FWD}, Fr\'{e}chet distance in wavelet-packet space (adapted from \citealt{veeramacheneni2025fwd}); \textbf{LPIPS} \citep{zhang2018lpips}, learned perceptual similarity (frozen VGG, CHMs normalized to $[0,1]$ and replicated to 3 channels).

\section{Results}
\label{sec:results}

All results are evaluated on the year-aligned test set (2,530 tiles) against LiDAR ground truth. Encoder and flow ablations use the 4,299-tile aligned training split. The AE/pixel controls use 4,299 tiles and 40 epochs; the temporal control uses 3,897 tiles and 10 epochs without DINOv3. The main AR-flow is trained on the full \textasciitilde197k pan-CONUS corpus (98 epochs, 8xA100).

\subsection{Headline: Sub-Meter CHM from Free Sentinel-2}
\label{sec:headline}

Table~\ref{tab:headline} compares both operating points with Meta CHMv2, the strongest commercial-input baseline in our evaluation. The Anchor trains for 20 epochs on \textasciitilde197k tiles; AR-flow trains for 98 epochs on the same corpus and uses a Dormand-Prince adaptive solver at inference. The Anchor emphasizes height accuracy, whereas AR-flow emphasizes perceptual fidelity.

\begin{table*}[t]
\centering
\caption{\textbf{Headline operating points.} Free S2+DINOv3 input vs commercial-input Meta CHMv2.}
\resizebox{\linewidth}{!}{
\begin{tabular}{lrrrrrrl}
\toprule
Operating point & MAE (m) & RMSE (m) & EE & SWD & FWD ($\times 10^3$) & LPIPS & Input \\
\midrule
Meta CHMv2 \citep{brandt2026meta} & 2.90 & 4.35 & 0.380 & 1.31 & 11.06 & 0.395 & 0.6~m Maxar \\
\textbf{Anchor} (S2+DINOv3, \textasciitilde197k tiles, 20 ep) & 2.83 & 4.06 & \textbf{0.326} & 1.13 & 3.47 & 0.436 & 10~m S2 \\
\textbf{AR-flow} (S2+DINOv3, \textasciitilde197k tiles, 98 ep) & 3.68 & 5.43 & 0.370 & \textbf{0.208} & \textbf{2.32} & \textbf{0.391} & 10~m S2 \\
\bottomrule
\end{tabular}}
\label{tab:headline}
\end{table*}

The Anchor achieves 2.83~m MAE vs Meta's 2.90~m from free 10~m input (paired bootstrap 95\% CI $[-0.10, -0.03]$~m, $n{=}2{,}530$). This cohort average masks regional differences: the Anchor has lower MAE in 27/69 ecoregions, with gains in tall-canopy forests and losses in arid and sparse-canopy regions (Section~\ref{sec:eco_analysis}). AR-flow improves all three perception metrics relative to Meta, with paired-bootstrap 95\% CIs excluding zero: SWD $\Delta{=}{-}0.90$ [$-1.09$, $-0.70$], FWD $\Delta{=}{-}6{,}481$ [$-9{,}098$, $-5{,}508$], LPIPS $\Delta{=}{-}0.005$ [$-0.007$, $-0.003$], trading 0.85~m MAE relative to the Anchor for sharper crowns.

\textbf{Autoencoder reconstruction benchmark.} The CHM autoencoder reaches 0.79~m MAE on the LiDAR encode-decode roundtrip (Appendix~\ref{sec:ceiling_appendix}). This measures reconstruction error, not a strict lower bound on every prediction decoded from the latent.

\subsection{Conditioning Encoder Ablation (Matched Compute)}
\label{sec:encoder_ablation}

We compare seven conditioning configurations while holding compute fixed (Stage~1, pure MSE, 10 epochs, 4,299 training tiles; Appendix~\ref{sec:encoder_appendix}). S2-only (3.376~m MAE) outperforms every foundation-model-only arm (3.582--4.384~m), and every S2+FM combination improves on S2-only. S2+AlphaEarth has the lowest MAE, 3.030~m, which is 0.23~m below S2+DINOv3. We retain S2+DINOv3 because its public checkpoint matches Meta CHMv2 and supports seasonal inputs; AlphaEarth embeddings are annual. The AlphaEarth gain and the temporal gain in Section~\ref{sec:temporal_ablation} come from separate experiments and should not be added. Sharing DINOv3 controls the checkpoint choice, but inputs, training data and downstream recipes still differ between pipelines.

\subsection{Stage-2 Flow Ablation: Residual vs Direct Target (Matched Compute)}
\label{sec:flow_ablation}

Both flow arms use identical backbone and conditioning; the only difference is the target: AR-flow on $r = z_y - \mu(c)$ vs plain flow on $z_y$ directly. At matched compute the AR-flow reaches 4.10~m MAE and FWD 2.40 against the plain flow's 4.67~m and 2.47, with the full-training Stage~1 reference at 2.83~m (Table~\ref{tab:transport}, Appendix~\ref{sec:flow_appendix}).

The full-training Anchor provides a reference for the perception-distortion trade-off \citep{blau2018perception}; it is not a matched ablation arm. Between the two matched flows, direct flow has slightly lower SWD, but AR-flow has lower FWD and about 0.58~m lower MAE. The anchored pipeline also retains the deterministic output as a separate operating point. At full scale, AR-flow is within 0.85~m MAE of the Anchor (Table~\ref{tab:headline}). The Anchor supplies an estimate of mean canopy structure, and the flow models the remaining latent variation.

\subsection{Does the Complexity Earn Its Place? (Matched Compute)}
\label{sec:complexity_ablation}

Two further controls isolate the domain-specific autoencoder and the latent parameterisation, both trained end to end for 40 epochs on 4,299 tiles rather than compared on reconstruction benchmarks. \textbf{Domain-specific autoencoder.} Replacing our domain CHM autoencoder with a stock ImageNet DC-AE at identical latent geometry ($f{=}64$, 128 channels, so Stage~2 receives the same 225 tokens) costs 0.22~m of Stage-1 MAE, 3.23 against 3.45, despite the stock model carrying 677~M parameters to our 41~M. The stock decoder also represents less bare ground: only 11.3\% of its pixels fall below 0.5~m, compared with our 19.3\% and LiDAR's 37.5\%. \textbf{Latent versus pixel space is close to a tie.} A pixel-space head on the same conditioning, split, epochs and batch reaches 3.28~m against the latent arm's 3.23~m, a gap four times smaller than the autoencoder effect (Appendix~\ref{sec:complexity_supp}).

The latent is justified primarily by the compact representation used by Stage~2; Appendix~\ref{sec:latent_rationale} discusses this computational trade-off.

\subsection{Temporal Information (Matched Compute)}
\label{sec:temporal_ablation}

A single three-month composite supports quarterly refresh but omits phenological information from other seasons. We measure the value of that information with a controlled temporal ablation. Two Stage-1 anchors were trained identically apart from their Sentinel-2 input: arm A replicates one seasonal composite four times, arm B concatenates four real seasonal composites (DJF/MAM/JJA/SON). Replication equalises input width and capacity, so the only variable is whether those channels carry four seasons or four copies of one. Architecture, parameter count, frozen autoencoders, seed, training tiles, augmentation and schedule are held fixed.

The four-season stack lowers MAE by \textbf{0.219~m} (95\% CI $[0.160, 0.283]$) and vegetation MAE by 0.307~m $[0.236, 0.382]$; intervals come from a paired bootstrap over whole EPA L3 ecoregions rather than tiles, since tiles cluster spatially, and the stack wins 10,000 of 10,000 resamples (Table~\ref{tab:temporal}, Appendix~\ref{sec:temporal_appendix}). The gain extends across ecoregions: of the 41 ecoregions with at least 20 tiles, 38 favour the stack and the three that do not sit within 0.02~m of zero. This pattern is consistent with phenological information: moist forested regions gain most (Puget Lowland 1.20~m), while arid basins gain little. These matched-compute arms train from scratch for 10 epochs without DINOv3, so only the delta is comparable, not the absolute values.

The single seasonal composite therefore trades 0.219~m MAE for quarterly cadence. Appendix~\ref{sec:temporal_appendix} discusses the relationship to seasonal substitution and time-series methods.

\subsection{Per-Ecoregion Analysis}
\label{sec:eco_analysis}

Similar cohort-level MAE (Table~\ref{tab:headline}) masks regional variation. We group the 69 test ecoregions by Meta CHMv2's per-ecoregion MAE (Appendix~\ref{sec:forest_plot}). These are error-defined groups, not direct canopy-height classes.

The Anchor has lower MAE in 18 of 33 regions in the highest-MAE group, but Meta leads in 42 of 69 regions overall. Giving each region equal weight yields 3.32~m MAE for the Anchor and 3.27~m for Meta, reversing the tile-average ordering. AR-flow improves on Meta in SWD in 61/69 regions and FWD in 56/67.

\subsection{SOTA Comparison}
\label{sec:sota}

We compare both operating points against eight baselines spanning commercial (Meta CHMv2), free multispectral (ETH Global, GLAD), SAR-fusion (EchoSat, \citealt{pauls2024canopy}) and spaceborne LiDAR (GEDI, Simard) sources, all bilinearly resampled to 0.5~m on the 2,530 aligned test tiles.

\begin{table*}[t]
\centering
\caption{SOTA comparison on 2,530 aligned test tiles. All metrics use a common 0.5~m evaluation grid; native CHM grid reports output spacing. LPIPS is reported only for our outputs and Meta; $^{\dagger}$out of vintage by construction: \citet{simard2011canopy} is built on 2005 ICESat/GLAS and ancillary maps, roughly fifteen years older than any subset we can vintage-match, and is listed for landscape completeness rather than as a like-for-like comparison. Dashes indicate metrics not reported for a product.}
\resizebox{\textwidth}{!}{
\begin{tabular}{llcrrrrrr}
\toprule
Product & Input (pixel size) & Native CHM grid & MAE (m) & RMSE (m) & EE & SWD & FWD ($\times 10^3$) & LPIPS \\
\midrule
Meta CHMv2 \citep{brandt2026meta} & Maxar Vivid2 (0.6~m) & 1~m & 2.90 & 4.35 & 0.380 & 1.31 & 11.06 & 0.395 \\
\textbf{Ours Anchor} & S2 (10~m) + DINOv3 & 0.5~m & \textbf{2.83} & 4.06 & \textbf{0.326} & 1.13 & 3.47 & 0.436 \\
\textbf{Ours AR-flow} & S2 (10~m) + DINOv3 & 0.5~m & 3.68 & 5.43 & 0.370 & \textbf{0.208} & \textbf{2.32} & \textbf{0.391} \\
GLAD \citep{potapov2021glad} & Landsat (30~m) & 30~m & 4.08 & 5.48 & 0.613 & 2.13 & 19.17 & -- \\
LandFire FCH \citep{landfire2023} & Landsat composite (30~m) & 30~m & 4.70 & 6.30 & 0.563 & 2.58 & 21.83 & -- \\
GEDI RH98 \citep{dubayah2020gedi} & spaceborne LiDAR & sparse & 5.65 & 7.64 & 0.422 & 4.19 & 27.55 & -- \\
EchoSat \citep{pauls2026echosat} & S1+S2+PALSAR-2+DEM & 10~m & 6.22 & 7.43 & 0.421 & 4.87 & 18.63 & -- \\
\citet{pauls2024canopy} & S1+S2 (10~m) & 10~m & 6.65 & 7.55 & 0.445 & 4.95 & 18.77 & -- \\
ETH Global \citep{lang2023high} & S2 (10~m) & 10~m & 7.05 & 8.47 & 0.525 & 5.46 & 15.47 & -- \\
\citet{simard2011canopy}$^{\dagger}$ & spaceborne LiDAR & 1~km & 9.25 & 10.45 & 0.909 & 6.59 & 28.54 & -- \\
\bottomrule
\end{tabular}}
\label{tab:sota}
\end{table*}

On this evaluation grid, the Anchor has the lowest MAE among the free-input products and is comparable to Meta; AR-flow has the lowest SWD and FWD. NEON ($n = 8$) gives similar MAE ($+0.02$~m) on the corresponding pixels, a small sample that cannot establish general robustness; GEDI agreement increases at coarse scales. Both are detailed in Appendix~\ref{sec:cross_lidar}. A common 2019--2020 LiDAR subset (Appendix~\ref{sec:vintage}) retains the distortion ranking within its re-scoring run, without exactly aligning every product vintage. Train-test MAE gaps of 0.01--0.06~m do not rule out spatial dependence or memorization. Per-pixel height-binned error, RAPSD, and height KDE analyses appear in Appendices~\ref{sec:height_bins_appendix}--\ref{sec:kde_appendix}.

\textbf{Resolution-matched comparison.} Table~\ref{tab:sota} scores every product on the common 0.5~m grid, which penalizes coarse products for unresolved fine-scale structure. Scored instead at each product's native cell (Appendix~\ref{sec:resmatched_appendix}, Table~\ref{tab:resmatched}), every baseline improves, and the Anchor remains lower within each paired comparison: our Anchor goes from 2.83~m to 1.37~m at 30~m against LandFire's 4.70 to 3.86. Aggregation removes sub-cell error from the fine-resolution outputs, so the wider margin depends on evaluation scale. Reporting both grids distinguishes this effect from performance at 0.5~m.

\subsection{Crown-Scale Structure}
\label{sec:crown}

Crown delineation and fuel mapping require per-tree structure that MSE-optimal CHMs suppress. We assess crown statistics and detection over 2,225 tiles and 2,395,143 LiDAR crowns, running \citet{silva2016itcd} ITCD uniformly on LiDAR, the Anchor, the AR-flow and Meta CHMv2. Detection depends on a matching threshold, whereas spectral and distributional measures do not. We report both because they capture different aspects of structure (Appendix~\ref{sec:metric_taxonomy}).

\textbf{Spectral structure.} A 16-tile radial-power probe against LiDAR (Appendix~\ref{sec:rapsd_appendix}) gives, at the 10~m input scale, retained power of 0.008 for the Anchor, 0.195 for Meta and 0.95 for the AR-flow, with 50\% crossover wavelengths of 83~m, 18.5~m and 4~m. Power measures spatial variation, not its correct placement. AR-flow's high retained power therefore does not establish recovery of real structure below the input pixel. The Anchor retains under 1\% of LiDAR's crown-scale variance in this probe, leaving little local variation for a crown detector to identify.

\begin{table*}[t]
\centering
\caption{\textbf{Crown-scale statistics} (2,225 tiles; Ripley over the 1,496-tile four-product intersection). The AR-flow is closest to LiDAR on height, tile-mean crown area and arrangement, and trails Meta on crown count.}
\begin{tabular}{lrrr}
\toprule
Statistic & Anchor & AR-flow & Meta CHMv2 \\
\midrule
Crown-count recovery (1.0 = LiDAR) & 0.21 & 0.78 & \textbf{0.82} \\
Crown-height KS vs full LiDAR field (lower better) & 0.171 & \textbf{0.059} & 0.152 \\
Tile-mean crown-area KS vs LiDAR (lower better) & -- & \textbf{0.096} & 0.205 \\
Ripley $L(r){-}r$ departure, $r \in [10,50]$~m (closer to 0) & $+5.64$ & $\mathbf{-1.40}$ & $+4.11$ \\
\bottomrule
\end{tabular}
\label{tab:crown_struct}
\end{table*}

\textbf{Spatial arrangement.} Matching crown counts and sizes does not ensure that crowns form realistic spatial patterns. Summarised over $r \in [10,50]$~m, beyond the detector's 3--6~m minimum-spacing floor, the AR-flow departs from LiDAR by $-1.40$ (95\% CI $[-2.44, -0.62]$, $n{=}1{,}496$ tiles), against $+4.11$ for Meta and $+5.64$ for the Anchor: 2.9$\times$ and 4.0$\times$ closer respectively, and closer at every radius in the band. The result is unchanged under density matching ($-1.41$), so it is not an artifact of the crown-count differences above. The interval excludes zero: AR-flow \emph{under}-clusters, reaching about 70\% of LiDAR's clustering amplitude. Stand-level clumping estimates derived from it are therefore biased toward regularity. Appendix~\ref{sec:metric_taxonomy} gives the conifer breakdown and explains how detector-imposed spacing affects the comparison.

\begin{table}[h]
\centering
\caption{\textbf{Individual-tree-crown detection}, \citet{silva2016itcd} ITCD at IoU $\geq 0.5$, pooled over 2,225 tiles. The commercial-input baseline leads; all three products sit near the floor. Appendix~\ref{sec:iou_sweep} gives the full IoU 0.2--0.5 sweep.}
\begin{tabular}{lrrr}
\toprule
Model & Recall & Precision & F1 \\
\midrule
Anchor & 0.013 & 0.061 & 0.021 \\
AR-flow & 0.037 & 0.047 & 0.041 \\
Meta CHMv2 & \textbf{0.062} & \textbf{0.076} & \textbf{0.068} \\
\bottomrule
\end{tabular}
\label{tab:crown}
\end{table}

\textbf{Individual-crown detection.} At IoU $\geq 0.5$, Meta leads with F1 0.068, compared with 0.041 for AR-flow from much coarser input imagery. All three sit near the floor, expected when crowns are sub-pixel: published IoU~$\geq 0.5$ scores come from centimetre-scale imagery, and reference provenance alone spans much of the range \citep{allen2025treecrowns}. Across IoU 0.2--0.5, AR-flow recall remains 2.4--3.0$\times$ the Anchor's, although absolute recall falls twelvefold (Appendix~\ref{sec:iou_sweep}). The relative gain therefore persists across the tested thresholds. AR-flow's lower matched-crown height error relative to Meta is confined to conifer stands (5.96 vs 6.20~m, $n{=}1{,}011$).

Sub-meter sampling alone does not yield sub-meter structure. The two operating points share resolution, conditioning, latent and decoder, and differ only in the loss; that difference is what separates a field with 0.8\% of LiDAR's crown-scale power from one with 95\%.

\subsection{Inference Efficiency and Seasonal Cadence}
\label{sec:seasonal}
\label{sec:efficiency}

Per-tile inference takes \textasciitilde3.15~s on A10G (\textasciitilde1.3~s on A100), dominated by the DINOv3 encoding and CHM autoencoder decode rather than the flow ODE itself; the AR-flow's residual parameterization yields a smoother velocity field, requiring 8\% fewer adaptive solver evaluations than the plain flow (63 vs 69). This efficiency makes continental-scale quarterly refresh tractable. 

To evaluate seasonal generalization, we hold the main Anchor fixed on 789 tiles and substitute four S2 composites (DJF, MAM, JJA, SON) per tile against the same LiDAR target. Appendix~\ref{sec:seasonal_fig_appendix} separately illustrates AR-flow predictions from the seasonal pilot.

MAM/JJA/SON stay within 1.32~m of each other, supporting three-quarter cadence, while DJF degrades by 1.89~m under deciduous leaf-off; the operational deliverable is therefore MAM/JJA/SON refresh. Per-quarter figures are in Appendix~\ref{sec:seasonal_table_appendix}.

\section{Discussion and Conclusion}
\label{sec:discussion}
\label{sec:conclusion}

We demonstrated sub-meter canopy-height super-resolution from free Sentinel-2, trained on 197k LiDAR tiles spanning 69 CONUS ecoregions. The shared conditioning supports two outputs for different forestry tasks. The Anchor has similar cohort-mean MAE to Meta CHMv2 from free 10~m input (Table~\ref{tab:headline}); AR-flow improves SWD, FWD and LPIPS at the cost of higher MAE.

The two outputs differ substantially in crown-scale structure despite sharing sampling, conditioning, latent space and decoder (Section~\ref{sec:crown}). In the 16-tile spectral probe, the Anchor retains under 1\% of LiDAR's power at the 10~m input scale, compared with 95\% for AR-flow. AR-flow also more closely matches LiDAR's crown-height, size and arrangement statistics. These improvements concern structural plausibility: Meta still leads individual-crown detection at IoU 0.5, and AR-flow's synthesized detail under-clusters crowns and lowers Block R$^2$. The distinction matters for use: the Anchor emphasizes mean-height accuracy, while AR-flow provides finer structure for fuel-mapping and silvicultural analyses, subject to the limits below.

\textbf{Operational implications.} Estimated inference costs are \$6--9K per forest-extent CONUS pass; quarterly global CHM refresh costs \textasciitilde\$300--460K/yr (\textasciitilde230K A100-hours; Appendix~\ref{sec:cost_appendix}), four orders of magnitude below airborne LiDAR at comparable extent. Such maps could help target new airborne acquisitions to areas of possible change. The architecture consumes a single composite, so shorter input windows offer a route toward weekly updates where cloud-free coverage permits. This would require training and validation with sparser observations; global deployment would also require LiDAR supervision and evaluation in underrepresented biomes.

\textbf{What the perception arm costs.} The AR-flow's injected high-frequency detail lowers pixel accuracy: cohort-mean Block R$^2$ is $+0.29$ for the Anchor, $-0.00$ for Meta and $-1.20$ for the AR-flow over 1,844 Anchor/AR-flow tiles and 1,847 Meta tiles, so its below-block variation is worse than predicting each block's own mean. This loss of spatial accuracy limits the use of generated detail. AR-flow also under-clusters crowns ($-1.40$, 95\% CI $[-2.44, -0.62]$, Section~\ref{sec:crown}), so clumping estimates lean toward regularity. Matched-crown error is a diagnostic rather than a per-tree measurement: with biomass scaling as \textasciitilde$h^{2.4}$, a 5--6~m crown-height error confines biomass claims to stand and landscape aggregates.

\textbf{Coverage and robustness.} DJF MAE rises \textasciitilde1.7x under deciduous leaf-off (Section~\ref{sec:seasonal}), so three-quarter cadence (MAM/JJA/SON) is the operational deliverable. 3DEP label noise includes flight-line seam artifacts (Appendix~\ref{sec:failure_appendix}), and NEON cross-validation ($\Delta{=}{+}0.02$~m, Appendix~\ref{sec:cross_lidar}) is encouraging but too small to establish cross-program robustness. All training is on CONUS: a zero-shot probe on four tiles each from Norway, British Columbia and New Zealand gives the largest errors in NZ's tall native forest, with different reference data and insufficient samples to establish regional transfer performance (Appendix~\ref{sec:ood_probe}). Tropical and mangrove biomes remain unevaluated, and the post-fire $\Delta$CHM alarm (Appendix~\ref{sec:disturbance}) flags disturbance without a calibrated biomass-loss estimate.

\textbf{Intermediate operating points.} Linear interpolation between the Anchor and AR-flow provides intermediate operating points (Appendix~\ref{sec:intermediate_appendix}); preliminary UQ shows per-pixel $\sigma$ correlates with error (Spearman $\rho{=}0.41$) but is not calibrated (65.7\% coverage at $2\sigma$), motivating calibrated intervals as future work.

\textbf{Future work.} Priorities include S1 SAR coherence conditioning for additional structural information, geographic expansion through pan-tropical LiDAR, and calibrated prediction intervals. We also plan accuracy and cost comparisons with canopy-height models based on National Agriculture Imagery Program (NAIP) imagery and canopy heights derived from NAIP photogrammetric point clouds. The planned release contains data rather than code (Appendix~\ref{sec:data_availability}); model weights remain institutional IP.

\textbf{Broader impact.} A free CHM sampled at 0.5~m could support stand- and landscape-level carbon accounting (REDD+ MRV), wildfire fuel mapping, and ecological monitoring. We do not release weights (institutional IP), limiting misuse potential.

\label{sec:mainend}

\section*{LLM Usage Statement}
\addcontentsline{toc}{section}{LLM Usage Statement}

Claude and Codex assisted with text editing and formatting of the manuscript.

{
    \small
    \bibliographystyle{ieeenat_fullname}
    \bibliography{references}

}

 \clearpage
\makeatletter\immediate\write\@auxout{\string\gdef\string\cvprsplitpage{\the\c@page}}\makeatother
\onecolumn
\raggedbottom 
\makeatletter\let\Hy@backout\@gobble\makeatother
\newcommand{\fitwidth}[1]{\resizebox{\ifdim\width>\linewidth\linewidth\else\width\fi}{!}{#1}}
\hypersetup{pageanchor=false}
\setcounter{page}{1}
\begin{center}
{\Large\textbf{\thetitle}}\\[0.5em]
{\large Supplementary Material}
\end{center}
\vspace{1em}
\appendix

\section{Metric and Experiment Conventions}

\label{sec:metric_conventions}

The primary benchmark reports height errors against airborne 3DEP on a common 0.5~m grid. MAE first averages absolute height differences over valid pixels within each tile, then averages tiles equally. Non-canopy pixels are included unless a vegetation mask is stated. A pixel-pooled MAE would weight tiles by valid area and is a different statistic. The regional macro-average instead gives each ecoregion equal weight. Distributional metrics such as SWD and FWD should not be reconstructed by averaging regional scores.

The model name alone does not identify an experiment. Table~\ref{tab:headline} uses the main models trained on approximately 197k tiles. The encoder ablation uses 4,299 training tiles and 10 epochs; the autoencoder and pixel-head controls use 4,299 tiles and 40 epochs; the temporal-stack ablation uses 3,897 tiles, 10 epochs, and no DINOv3. The residual/direct-flow comparison uses separate ablation runs, with the main Anchor repeated only as a reference. Absolute MAEs across these training regimes do not isolate any single design choice.

Independent re-scoring controls use the same scorer within each comparison. Native-grid results change the evaluation scale and jointly valid mask; vintage results change the cohort. Neither should replace the primary scores. The 789-tile seasonal analysis holds the main Anchor fixed and substitutes its input season; the 50-tile AR-flow illustration and the four-season training ablation are separate experiments. Cross-LiDAR and non-CONUS results additionally change the reference data. Crown detection uses pooled counts, whereas crown-size KS compares distributions of tile-mean areas.

\section{Design-Choice Ablations}
\label{sec:design_ablations}

\subsection{Stage-2 Flow Target Ablation}
\label{sec:flow_appendix}

\begin{table}[H]
\centering
\caption{\textbf{Flow target ablation.} The two flow arms use identical backbone, conditioning and compute. Stage~1 is the full-training reference, not a matched ablation arm.}
\begin{tabular}{lrrrr}
\toprule
Model & MAE (m) & EE & SWD & FWD ($\times 10^3$) \\
\midrule
Stage~1 reference (full training) & \textbf{2.83} & \textbf{0.326} & 1.13 & 3.47 \\
\textbf{AR-flow} (residual target $r = z_y - \mu(c)$) & 4.10 & 0.384 & 0.391 & \textbf{2.40} \\
Plain flow matching (direct target $z_y$) & 4.67 & 0.400 & \textbf{0.312} & 2.47 \\
\bottomrule
\end{tabular}
\label{tab:transport}
\end{table}

\subsection{Autoencoder and Latent-Space Controls}

 \label{sec:complexity_supp}

 \textbf{Domain versus stock autoencoder.} Both arms train for 40 epochs on the 4,299-tile aligned training subset with S2 and DINOv3 conditioning. These are separate runs from both the main 197k-tile models and the 10-epoch encoder ablation. We retrained the Stage-1 Anchor with the frozen target autoencoder replaced by a stock ImageNet DC-AE (\texttt{dc-ae-f64c128-in-1.0}) of identical latent geometry (\(f{=}64\), 128 channels), so both arms give Stage~2 the same 225 tokens. MAE rises from 3.23~m to 3.45~m (difference 0.22~m, 95\% CI \([0.19, 0.25]\)), although the stock model is over ten times larger (677M parameters). The stock decoder under-represents bare ground: 37.5\% of LiDAR pixels are below 0.5~m, against 19.3\% for our autoencoder and 11.3\% for the stock one, which pushes ground into the 1--8~m band. On vegetated pixels, the stock arm has slightly lower MAE (6.17 vs 6.26~m); these masked errors are separate from the all-valid-pixel results above.

 \textbf{Latent versus pixel space.} A pixel-space head on the same S2 and DINOv3 conditioning, split, epochs and batch reaches 3.28~m against the latent head's 3.23~m (difference 0.05~m, 95\% CI \([0.02, 0.08]\)). The small accuracy gap is empirical; latent-space and pixel-space L2 objectives need not yield the same decoded minimizer. Its role is Stage~2: the flow transports the residual in the anchor's latent, and sampling cost scales with tokens times solver steps. Our latent has 225 tokens, against 14,400 for an \(f{=}8\) SD-VAE latent and 921,600 for raw pixels.

 \subsection{Temporal Stack Ablation}
\label{sec:temporal_appendix}

This is distinct from the seasonal substitution in Section~\ref{sec:seasonal}, which scored a frozen Anchor on each quarter in turn and showed that choosing a good season matters. The present experiment shows what that could not: combining seasons improves on the matched replicated-single-season control. The single composite is therefore a deliberate trade with a measured price. A four-season stack requires annual updates, or imagery up to nine months stale in every map, and we buy quarterly cadence with 0.219~m. \citet{kalinicheva2025threasure} likewise credit phenology in a Sentinel-2 time series and spend it on annual maps over France; our ablation puts a number on the same signal in the opposite direction.

\begin{table}[H]
\centering
\caption{\textbf{Temporal stack ablation}, matched compute. Intervals are a paired bootstrap resampling whole EPA L3 ecoregions rather than tiles, since tiles cluster spatially; clustering roughly doubles the interval and the stack still wins 10,000 of 10,000 resamples.}
\begin{tabular}{lrrrl}
\toprule
MAE (m), lower better & A single & B four-season & $\Delta$ (A$-$B) & 95\% CI \\
\midrule
All pixels & 3.201 & \textbf{2.982} & 0.219 & $[0.160, 0.283]$ \\
Vegetation ($>2$~m) & 5.591 & \textbf{5.284} & 0.307 & $[0.236, 0.382]$ \\
\bottomrule
\end{tabular}
\label{tab:temporal}
\end{table}

\subsection{Autoencoder Reconstruction Benchmark}
\label{sec:ceiling_appendix}

\begin{table}[H]
\centering
\caption{\textbf{Autoencoder reconstruction benchmark vs headline operating points.}}
\fitwidth{
\begin{tabular}{lrrrrr}
\toprule
Operating point & MAE (m) & RMSE (m) & EE & SWD & FWD ($\times 10^3$) \\
\midrule
\textbf{AE roundtrip} (LiDAR $\to$ encode $\to$ decode) & \textbf{0.79} & \textbf{1.33} & \textbf{0.148} & \textbf{0.46} & \textbf{1.04} \\
Anchor & 2.83 & 4.06 & 0.326 & 1.13 & 3.47 \\
AR-flow & 3.68 & 5.43 & 0.370 & 0.21 & 2.32 \\
\bottomrule
\end{tabular}}
\label{tab:ceiling}
\end{table}

The 0.79~m roundtrip is a reconstruction benchmark, not a strict lower bound on all predictions decoded from this latent.

\subsection{Conditioning Encoder Ablation}
\label{sec:encoder_appendix}

\begin{table}[H]
\centering
\caption{\textbf{Stage-1 encoder ablation.} All arms at matched compute (10 epochs, 4,299 tiles).}
\begin{tabular}{lrr}
\toprule
Encoder & MAE (m) & EE \\
\midrule
\textbf{S2 + DINOv3} (ablation) & 3.259 & 0.476 \\
S2 + Clay v1.5 & 3.268 & 0.482 \\
S2 + AlphaEarth & \textbf{3.030} & 0.490 \\
S2 only (no foundation model) & 3.376 & 0.472 \\
DINOv3 only (no S2 latent) & 4.384 & 0.385 \\
Clay v1.5 only & 3.974 & 0.395 \\
AlphaEarth only & 3.582 & 0.391 \\
\bottomrule
\end{tabular}
\label{tab:encoder}
\end{table}

\subsection{Why the Flow Operates in the CHM Latent}
\label{sec:latent_rationale}

We therefore do not defend the latent on Stage-1 accuracy -- the near-tie is empirical, since latent-space and pixel-space L2 objectives need not have the same decoded minimizer. We defend it as the space Stage~2 operates in, on two grounds. The flow transports $r = z_y - \mu(c)$ away from the anchor's own output, so a pixel-space anchor would require a different coupling to Stage~2. And sampling cost scales as tokens times solver steps: our CHM latent is 225 tokens against 14{,}400 for an off-the-shelf $f{=}8$ SD-VAE latent and 921{,}600 for raw pixels, so each ODE step moves 64$\times$ fewer tokens than the standard latent choice and 4{,}096$\times$ fewer than pixel space. That is what makes \textasciitilde3~s/tile, and hence continental refresh, tractable at all (Appendix~\ref{sec:latent_rationale}).

Section~\ref{sec:complexity_ablation} shows the latent ties pixel space on Stage-1 accuracy. Two properties justify it at Stage~2.

\textbf{Compactness.} A flow is not one forward pass: sampling integrates an ODE and every step moves every token, so cost scales as tokens times steps. On the counts in Section~\ref{sec:complexity_ablation}, a pixel-space flow is not a variant we declined to run; at 4{,}096$\times$ the tokens per step it is a different regime, and even the standard $f{=}8$ latent is a 64$\times$ heavier one.

\textbf{Richness.} The flow's endpoint is an arbitrary latent point and must decode to something resembling canopy rather than blur. That is what an ImageNet autoencoder lacks: the stock DC-AE at the same latent shape cannot represent bare ground, so points in its latent decode to a distribution the task does not contain. Pixels are rich but not compact; the ImageNet latent is compact but not rich.

Both claims are falsifiable and untested. A pixel-space flow matching our perception metrics would refute compactness; a flow in the stock latent would refute richness. \citet{freirich2021theory} motivate coupling the conditional-mean estimate to a generative transport; their result does not establish that this particular latent or flow implementation is necessary or optimal.

\section{Full Architecture Specification}
\label{sec:architecture_appendix}

\begin{table}[H]
\centering
\caption{\textbf{System components.}}
\fitwidth{
\begin{tabular}{llrlll}
\toprule
Component & Architecture & Params & Input & Output & Frozen? \\
\midrule
S2 encoder & DC-AE \citep{chen2024dcae}, $f{=}8$ & \textasciitilde12M & 12-band S2, $48 \times 48 \to 128 \times 128$ & $32 \times 16 \times 16$ latent & yes \\
CHM autoencoder & DC-AE \citep{chen2024dcae}, $f{=}64$ & \textasciitilde50M & 1-ch CHM, $960 \times 960$ & $128 \times 15 \times 15$ latent & yes \\
DINOv3 & ViT-L/16 SAT-493M \citep{simeoni2025dinov3} & \textasciitilde300M & S2 RGB (B04/B03/B02), $224 \times 224$ & 196 tokens, dim 1024 & yes \\
Stage~1 (Anchor) & UViT \citep{bao2023uvit} & \textasciitilde110M & S2 latent + DINOv3 x-attn & $128 \times 15 \times 15$ & frozen for Stage~2 \\
Stage~2 (AR-flow) & UViT \citep{bao2023uvit} & \textasciitilde110M & $(x_t, t)$ + DINOv3 x-attn & $128 \times 15 \times 15$ & no \\
\bottomrule
\end{tabular}}
\label{tab:architecture}
\end{table}

Both stages use the same UViT-style backbone \citep{bao2023uvit} (embed 768, depth 12, 12 heads) with added DINOv3 cross-attention at three decoder depths.

\subsection{The two autoencoders are not interchangeable}

Both follow DC-AE \citep{chen2024dcae} and both are trained from scratch on domain data with an L1 + LPIPS + adversarial objective, then frozen. They differ in role and in compression factor, and conflating them is the single easiest way to misread the system.

The \textbf{S2 source encoder} is a side channel. Its stage depths are $[3,5,3,3]$ with widths $[80,160,320,640]$; with a non-zero leading stage this gives $2^3 = 8\times$ spatial downsampling and 32 latent channels. The model input is a $48 \times 48$ pixel S2 crop (480~m at 10~m GSD, exactly the 960~px CHM footprint at 0.5~m), bilinearly upsampled to $128 \times 128$ before encoding, yielding a $32 \times 16 \times 16$ latent.

The \textbf{CHM target autoencoder} is the generative bottleneck the flow actually operates in. Its stage depths are $[0,2,3,1,1,1,1]$; the leading zero contributes a factor of 2 and five further non-terminal stages contribute $2^5$, giving $64\times$ downsampling with 128 latent channels. A $960 \times 960$ CHM crop becomes a $128 \times 15 \times 15$ latent: 225 spatial tokens, $64^2 = 4{,}096\times$ fewer positions than the pixel grid.

Quoting $f{=}8$ as ``the model's compression'' is therefore wrong; that is the conditioning path. The number that governs Stage~2's cost and expressiveness is $f{=}64$.

\subsection{Conditioning}

DINOv3 ViT-L/16 SAT-493M is used frozen and off the shelf, with no patch-embedding inflation. Its patch embedding is three-channel, so the stream sees only B04/B03/B02 resized to $224 \times 224$, giving $(224/16)^2 = 196$ patch tokens of dimension 1024. The red-edge, NIR and SWIR bands reach the model exclusively through the 12-band S2 autoencoder path. Normalisation uses the SAT-493M statistics, mean $(0.430, 0.411, 0.296)$ and standard deviation $(0.213, 0.156, 0.143)$.

\subsection{Training and inference, in pseudocode}

Prose is ambiguous about which tensor enters which stage, so both loops are stated directly. $E_{\text{S2}}$ and $E_{\text{CHM}}$ are the frozen encoders, $D_{\text{CHM}}$ the frozen CHM decoder, $\mu_\theta$ the Stage-1 UViT and $v_\phi$ the Stage-2 velocity head.

\vspace{0.4em}
\noindent\rule{\linewidth}{0.4pt}
{\small
\textbf{Stage 1 (Anchor), one step.} Given an S2 crop $s \in \mathbb{R}^{12 \times 48 \times 48}$ and CHM target $y \in \mathbb{R}^{1 \times 960 \times 960}$:
\begin{enumerate}\itemsep0em
\item $c_{\text{ae}} \leftarrow E_{\text{S2}}(\text{bilinear}(s, 128 \times 128))$ \hfill $\triangleright\ 32 \times 16 \times 16$
\item $c_{\text{vit}} \leftarrow \text{DINOv3}(\text{resize}(s_{\text{RGB}}, 224 \times 224))$ \hfill $\triangleright\ 196 \times 1024$
\item $z_y \leftarrow E_{\text{CHM}}(y)$ \hfill $\triangleright\ 128 \times 15 \times 15$
\item $\mathcal{L} \leftarrow \lVert \mu_\theta(c_{\text{ae}}, c_{\text{vit}}) - z_y \rVert_2^2$ \hfill $\triangleright$ pure L2, no perceptual term
\end{enumerate}

\textbf{Stage 2 (AR-flow), one step.} $\mu_\theta$ is frozen throughout:
\begin{enumerate}\itemsep0em
\item $r \leftarrow z_y - \mu_\theta(c)$ \hfill $\triangleright$ residual target, not $z_y$
\item $\varepsilon \sim \mathcal{N}(0, I)$, \quad $t \sim \mathcal{U}[0,1]$
\item $x_t \leftarrow (1-t)\,\varepsilon + t\,r$ \hfill $\triangleright$ linear path
\item $\mathcal{L} \leftarrow \lVert v_\phi(x_t, t, c) - (r - \varepsilon) \rVert_2^2$
\end{enumerate}

\textbf{Inference.} Integrate $dx/dt = v_\phi(x, t, c)$ from $\varepsilon$ at $t{=}0$ to $\hat r$ at $t{=}1$ with an adaptive solver \citep{dormand1980dopri}, then $\hat y = D_{\text{CHM}}(\mu_\theta(c) + \hat r)$. The anchor enters Stage~2 only through the target $r$; it is never concatenated to the velocity head's input.
}
\noindent\rule{\linewidth}{0.4pt}
\vspace{0.4em}

\subsection{Corpus and schedules}

The hero Anchor trains for 20 epochs and the hero AR-flow for 98 epochs on the \textasciitilde197k-tile corpus. The encoder ablation trains Stage~1 for 10 epochs on the 4,299-tile year-aligned subset; the autoencoder and pixel-head controls use the same training cohort for 40 epochs, varying one factor at a time; the temporal ablation of Section~\ref{sec:temporal_ablation} uses 3,897 training tiles, 10 epochs and seed 42, without DINOv3, which is why its absolute MAE is not commensurate with Table~\ref{tab:headline} and its absolute values are interpreted only within that control. Every transport and encoder arm is compared at identical epoch count.

Optimisation is shared across every run and is given in Section~\ref{sec:training}: AdamW at $1 \times 10^{-4}$, weight decay 0.03, cosine schedule with 50\% warmup to a $10^{-7}$ floor, gradient clipping at 1.0, and global batch size 16 in full precision. Sampling uses a Dormand-Prince adaptive solver \citep{dormand1980dopri} at inference and forward Euler ($N{=}32$) for training-time validation.

Sampling is seeded and fixed at inference (Section~\ref{sec:training}), so the AR-flow results are reproducible draws rather than a distribution over runs.

\section{Positioning Relative to Prior Work}
\label{sec:positioning_appendix}

\begin{table}[H]
\centering
\fitwidth{
\begin{tabular}{llll}
\toprule
Paper & Cross-modal & Frozen MSE Stage-1 & Stage-2 transport \\
\midrule
CorrDiff \citep{mardani2024corrdiff} & weak & yes & EDM-diffusion noise $\to$ residual, $\mu$-concat \\
ELIR \citep{cohen2025elir} & no & co-trained L2 & consistency flow: noised L2 estimate $\to$ clean (Tiny AE) \\
\citet{kim2025cbbdm} & yes (SAR$\to$optical) & no, source-cond & Brownian-bridge diffusion \\
C-DiffSET \citep{do2026cdiffset} & yes (SAR$\to$optical) & no & noise $\to$ y diffusion \\
\textbf{Ours} & yes (optical$\to$height) & yes & AR-flow: linear-path noise $\to$ residual, no $\mu$-concat, CHM autoencoder latent \\
\bottomrule
\end{tabular}}
\caption{Positioning relative to the closest prior work.}
\label{tab:positioning}
\end{table}

\section{Resolution-Matched Comparison: Method and Controls}
\label{sec:resmatched_appendix}

\begin{table}[H]
\centering
\caption{\textbf{Resolution-matched comparison} (MAE in m). Each product is scored at its native cell size, with our predictions aggregated to the same cells. The independent 0.5~m control agrees with Table~\ref{tab:sota} within 0.01~m. Counts reflect jointly valid tiles.}
\begin{tabular}{llrrrr}
\toprule
Product & Grid & $n$ tiles & Product MAE & Anchor MAE & AR-flow MAE \\
\midrule
\emph{(control)} Meta CHMv2 & 0.5~m & 2,530 & 2.892 & \textbf{2.825} & 3.673 \\
Meta CHMv2 & 1~m & 2,530 & 2.816 & \textbf{2.712} & 3.580 \\
ETH Global & 10~m & 2,527 & 6.845 & \textbf{1.838} & 2.440 \\
Pauls / AI4Forest & 10~m & 2,530 & 6.079 & \textbf{1.839} & 2.444 \\
EchoSat & 10~m & 2,530 & 5.880 & \textbf{1.839} & 2.444 \\
GLAD & 30~m & 2,526 & 3.102 & \textbf{1.367} & 1.714 \\
LandFire FCH & 30~m & 2,530 & 3.855 & \textbf{1.372} & 1.720 \\
\bottomrule
\end{tabular}
\label{tab:resmatched}
\end{table}

Table~\ref{tab:resmatched} scores each product at its own native cell size. Both the prediction and the LiDAR reference are aggregated to that cell by a plain block mean over jointly-valid pixels; cells with no valid pixel are dropped. Nodata is excluded rather than zero-filled, which matters: zero-filling scores the 20~px prediction margin as 0~m canopy against real LiDAR and inflates Anchor MAE by 0.30~m.

The 0.5~m row is a control. At that grid the pipeline reproduces Table~\ref{tab:sota} for all six baselines (Meta 2.892 vs 2.90, GLAD 4.081 vs 4.08, LandFire 4.703 vs 4.70, EchoSat 6.217 vs 6.22, AI4Forest 6.645 vs 6.65, ETH 7.050 vs 7.05) and for both our arms (Anchor 2.825 vs 2.83, AR-flow 3.673 vs 3.68). Any rescoring must pass that control before its other rows are trusted.

Scoring on the common 0.5~m grid penalises coarse products for sub-pixel structure they never claim to resolve; evaluating each at its native cell removes that penalty, and LandFire improves from 4.70~m to 3.86~m, Pauls from 6.65~m to 6.08~m. The mechanism cuts both ways and we state it plainly rather than let a reader derive it. Aggregation removes error at scales finer than the cell, so a 0.5~m prediction has far more error available to remove than a 30~m product that never resolved those scales: our Anchor improves from 2.83~m to 1.37~m between the 0.5~m and 30~m grids while LandFire, already at 30~m, gains only 0.85~m. The widened margin is therefore a property of the evaluation scale, not evidence of a larger capability gap than Table~\ref{tab:sota} reports.

Two products are excluded. \citet{simard2011canopy} is nominally $\sim$1~km, at which a 480~m tile degenerates to a single cell. GEDI is sparse footprints rather than a wall-to-wall raster and serves as the independent reference in Appendix~\ref{sec:cross_lidar}, so scoring it as a competitor would conflate the two roles.

\section{Seasonal Cadence: Per-Quarter Figures}
\label{sec:seasonal_table_appendix}

\begin{table}[H]
\centering
\caption{Seasonal cadence evaluation. The frozen Anchor scored on each quarterly composite in turn against the same LiDAR target.}
\fitwidth{
\begin{tabular}{lrrrrr}
\toprule
Quarter & $n$ & MAE (m) & RMSE (m) & EE & $\Delta$ MAE vs JJA \\
\midrule
\textbf{JJA (dominant acquisition season)} & 789 & \textbf{2.75} & 4.13 & 0.321 & -- \\
MAM (spring) & 789 & 4.07 & 5.61 & 0.323 & $+1.32$ \\
SON (autumn) & 789 & 3.99 & 5.39 & 0.324 & $+1.24$ \\
DJF (winter) & 789 & 4.64 & 6.28 & 0.326 & $+1.89$ \\
\bottomrule
\end{tabular}}
\label{tab:seasonal}
\end{table}

\section{Height-Binned Error Analysis}
\label{sec:height_bins_appendix}

\begin{figure}[H]
\centering
\includegraphics[width=\linewidth]{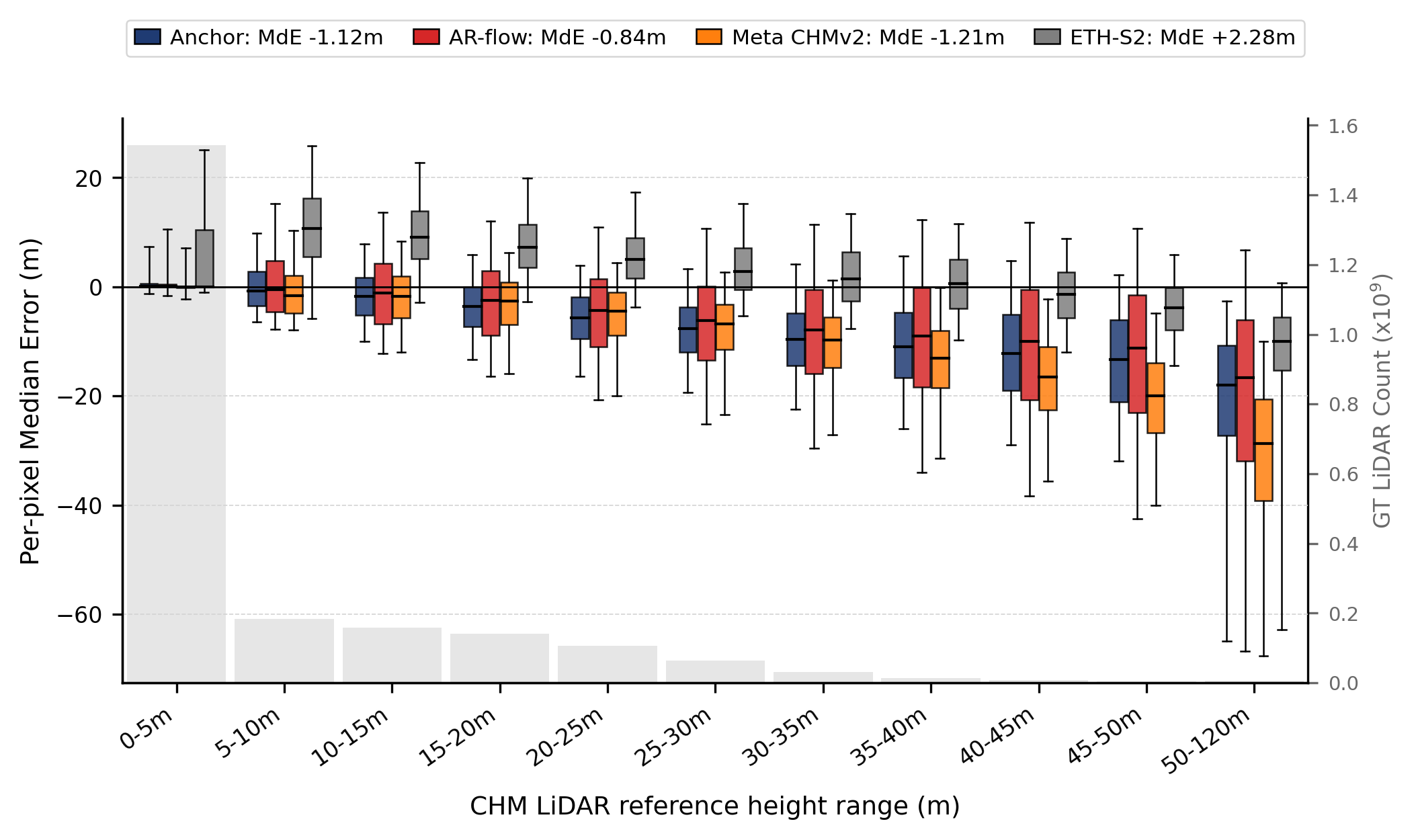}
\caption{\textbf{Per-pixel error binned by LiDAR reference height.} Boxes show interquartile range (25--75 percentile), whiskers the 5--95th percentile; grey bars give relative LiDAR pixel count per bin (right Y-axis, $\times 10^9$). Anchor and AR-flow from free 10~m input vs Meta CHMv2 (commercial Maxar) and ETH-S2 (free S2 comparator). All 2,530 tiles, no sub-sampling.  Errors are signed ( prediction minus LiDAR), and  the legend reports median signed error ( MdE) , not MAE. All products tend to  underestimate the  tallest canopy bins.}
\label{fig:height-bins}
\end{figure}

\section{Radially Averaged Power Spectral Density}
\label{sec:rapsd_appendix}

\begin{figure}[H]
\centering
\includegraphics[width=\linewidth]{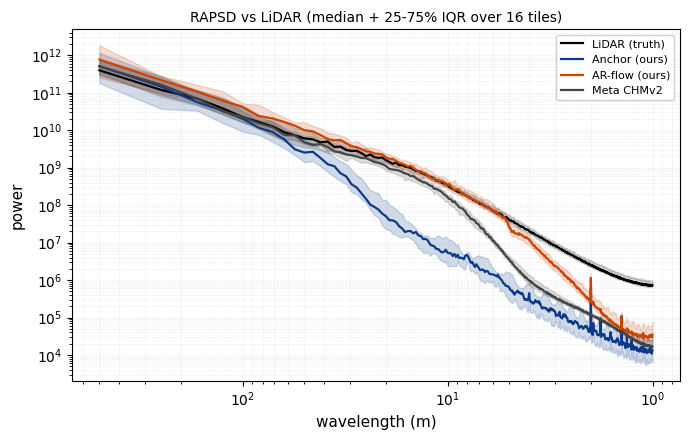}
\caption{\textbf{RAPSD vs LiDAR.} x: wavelength (m), y: power; curves LiDAR / Anchor / AR-flow / Meta CHMv2 with 25--75\% inter-tile band over 16 tiles.}
\label{fig:psd}
\end{figure}

At the 10~m Sentinel-2 input scale the Anchor retains 0.008 of LiDAR's power, Meta CHMv2 0.195 and the AR-flow 0.95; the wavelengths at which each curve falls to half of LiDAR's power are 83~m, 18.5~m and 4~m respectively. The crossover is a 50\% crossing of a smooth decay curve, so its exact value moves with the threshold and we report it for completeness rather than as a resolution figure. The threshold-free reading is the retained-power column, and its load-bearing entry is the Anchor's: a field holding under 1\% of LiDAR's variance at the input scale cannot support crown-scale inference at any output sampling.

\section{Crown-Scale Structure: Supporting Analyses}
\label{sec:crown_appendix_group}

\subsection{Full IoU Sweep}
\label{sec:iou_sweep}

Table~\ref{tab:crown} reports IoU $\geq 0.5$ only. The full sweep is below, pooled over the same 2,225 tiles and 2,395,143 LiDAR crowns, with a single matching definition throughout. Meta CHMv2 leads recall at every threshold. The quantity that is stable across thresholds is the Anchor-to-AR-flow recall ratio, 2.4$\times$ to 3.0$\times$, while absolute recall falls roughly twelvefold from IoU 0.2 to 0.5 -- which is why a single threshold cannot settle the question and why Section~\ref{sec:crown} rests on threshold-free measures.

\begin{table}[H]
\centering
\caption{\textbf{ITCD sweep}, \citet{silva2016itcd}, pooled recall / precision / F1 at four matching thresholds.}
\begin{tabular}{lrrrrrrrrr}
\toprule
& \multicolumn{3}{c}{Anchor} & \multicolumn{3}{c}{AR-flow} & \multicolumn{3}{c}{Meta CHMv2} \\
\cmidrule(lr){2-4}\cmidrule(lr){5-7}\cmidrule(lr){8-10}
IoU & R & P & F1 & R & P & F1 & R & P & F1 \\
\midrule
0.2 & 0.178 & 0.689 & 0.283 & 0.426 & 0.480 & 0.452 & \textbf{0.512} & 0.566 & \textbf{0.537} \\
0.3 & 0.085 & 0.401 & 0.141 & 0.236 & 0.296 & 0.262 & \textbf{0.321} & 0.385 & \textbf{0.350} \\
0.4 & 0.038 & 0.180 & 0.063 & 0.109 & 0.137 & 0.121 & \textbf{0.163} & 0.196 & \textbf{0.178} \\
0.5 & 0.013 & 0.061 & 0.021 & 0.037 & 0.047 & 0.041 & \textbf{0.062} & \textbf{0.076} & \textbf{0.068} \\
\bottomrule
\end{tabular}
\label{tab:iou_sweep}
\end{table}

\subsection{What Each Structural Measure Captures}
\label{sec:metric_taxonomy}

Section~\ref{sec:crown} uses several measures because each is blind to something the others catch. Reporting any one alone would overstate the case.

\begin{table}[H]
\centering
\small
\caption{\textbf{Structural measures and their blind spots.} Each answers a different question about the predicted height field; none is sufficient alone.}
\begin{tabular}{p{0.20\linewidth}p{0.26\linewidth}p{0.42\linewidth}}
\toprule
Measure & Question it answers & What it cannot see \\
\midrule
RAPSD & Is variance present at crown scale? & Placement. A field with correct texture in the wrong locations scores perfectly. \\
Crown-height / crown-size KS & Are the marginal distributions right? & Placement, and any joint structure between size and location. \\
Ripley $L(r){-}r$ & Is the spatial arrangement right? & Height. Two fields with identical point patterns may have unrelated canopy heights. \\
SWD / FWD / LPIPS & Does the field match reference patch statistics or features? & Geometric correctness; perceptual agreement does not guarantee accurate crown placement. \\
ITCD recall / precision & Can discrete objects be recovered? & Everything below the matching threshold, and the threshold is a free parameter. \\
Block R$^2$ & Is fine-scale pixel variation right? & Perceptual plausibility; it penalises correct-looking detail placed imperfectly. \\
\bottomrule
\end{tabular}
\label{tab:metric_taxonomy}
\end{table}

The AR-flow improves on the Anchor in spectral, distributional and detection measures, but trails it on Block R$^2$. It improves on the commercial-input baseline in spectral and distributional measures, trails it on object detection, and pays for the difference on Block R$^2$ (Appendix~\ref{sec:block_r2}). That pattern is what the perception-distortion trade-off predicts, and it is why the paper ships two operating points rather than choosing one.

Two details from the Ripley analysis in Section~\ref{sec:crown}. The AR-flow's under-clustering is concentrated in conifer stands ($-2.19$, $n{=}844$) against $-0.35$ in deciduous. And a caveat that runs in our favour: the detector enforces a 3--6~m minimum spacing, and that constraint binds on 24.1\% of Meta CHMv2's detections against 0.06\% of LiDAR's and 0.00\% of the AR-flow's. A coarser raster resampled onto a 0.5~m grid carries plateaus of tied local maxima, so part of Meta's crown placement is detector-imposed rather than canopy-driven. We have not adjusted the comparison for this and report it because it affects how the crown-count and arrangement rows should be read.

\section{Height Distribution (KDE)}
\label{sec:kde_appendix}

\begin{figure}[H]
\centering
\includegraphics[width=\linewidth]{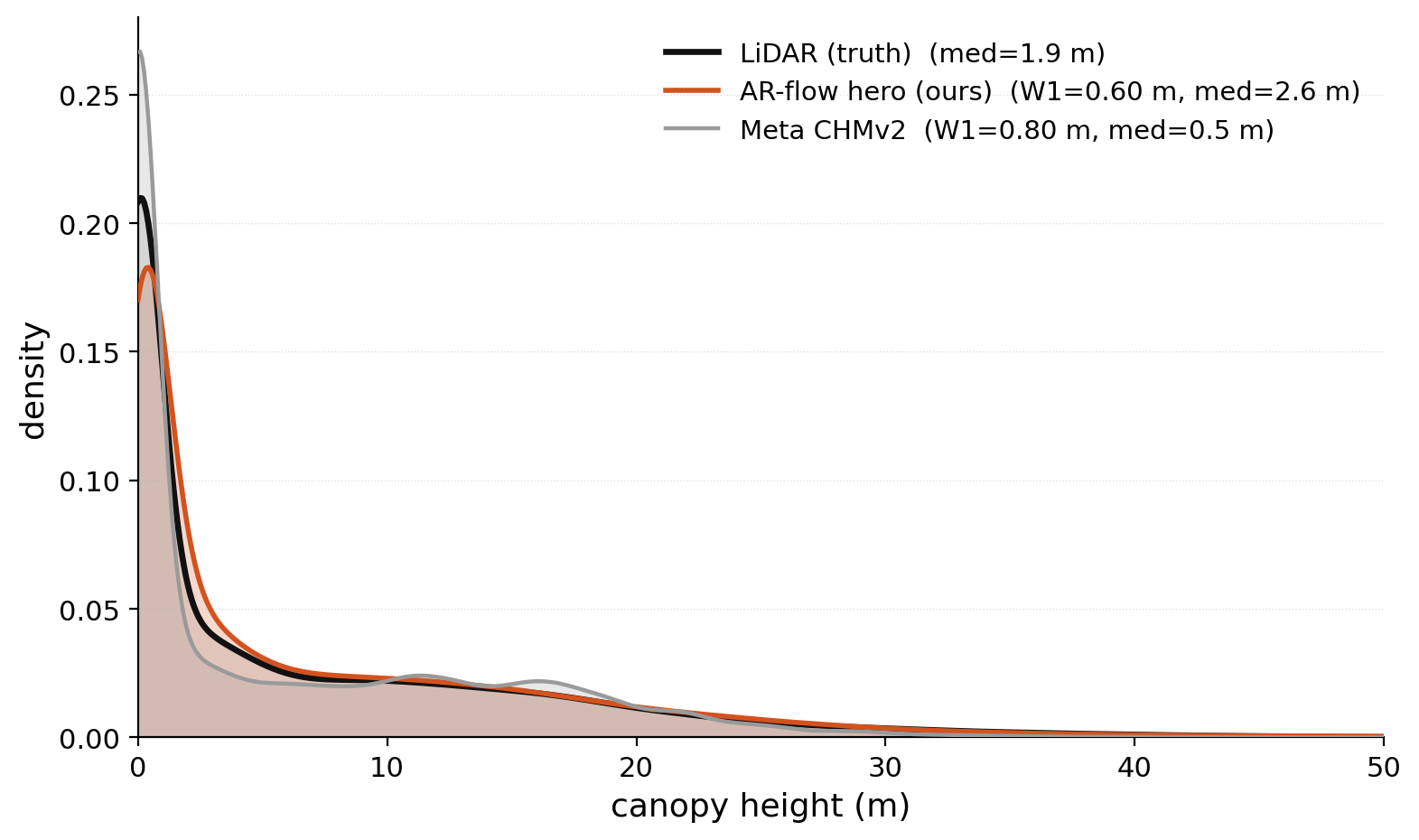}
\caption{ \textbf{Height distributions.} Curves  show LiDAR, AR-flow, and Meta CHMv2. The legend reports each median and the 1-Wasserstein distance to LiDAR in meters. This one-dimensional height-distribution distance is distinct from the patch-based SWD in Table~\ref{tab:headline}.}
\label{fig:kde}
\end{figure}

\section{Crown Detection Visual}

\label{sec:crown_appendix}

\begin{figure}[H]
\centering
\includegraphics[width=\linewidth]{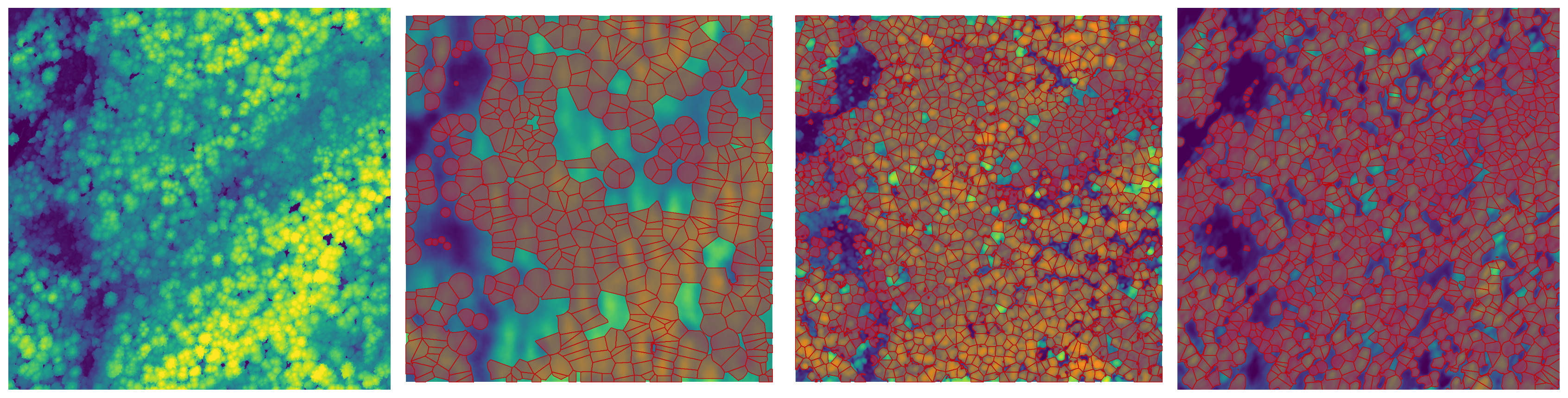}
\caption{\textbf{Individual-tree-crown detection across CHM sources} (tile \texttt{8327\_11991}, Coast Range). Panels left-to-right: LiDAR truth, Anchor, AR-flow, Meta CHMv2. Red fill marks predicted crowns matching a LiDAR crown at IoU $\geq 0.2$ (\citealt{silva2016itcd} ITCD, conifer parameters). Counts: LiDAR 1,549 crowns; Anchor 601 (491 matched ); AR-flow 1,296 (1,098 matched ); Meta 1,057 (945 matched ).}
\label{fig:crown_matched}
\end{figure}

\section{Seasonal Cadence Visual}
\label{sec:seasonal_fig_appendix}

\begin{figure}[H]
\centering
\includegraphics[width=\linewidth]{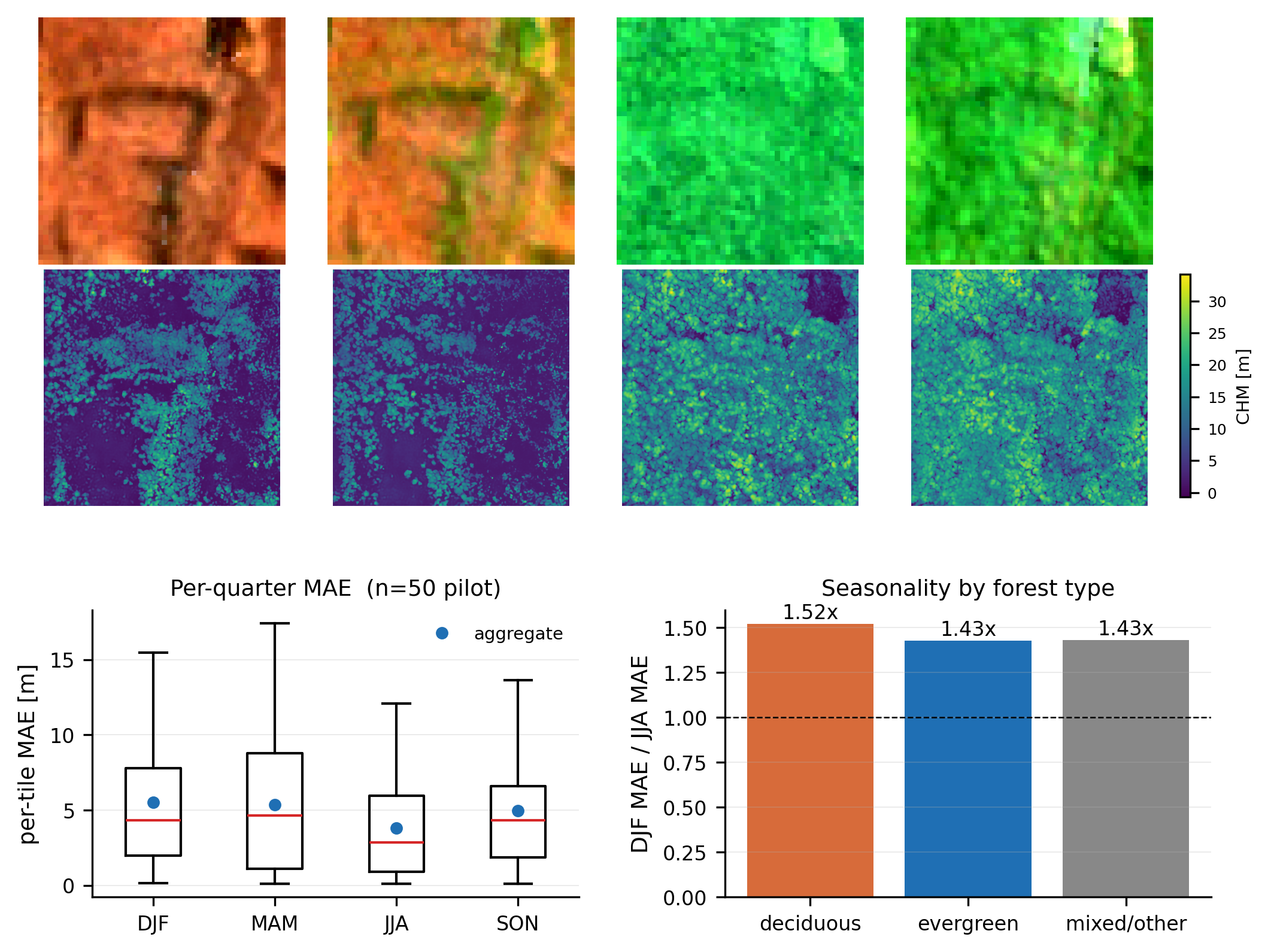}
\caption{\textbf{Seasonal cadence on a representative deciduous tile.} Interior Plateau, tile \texttt{6358\_18536}. Top row: per-quarter S2 RGB composites showing leaf-off (DJF) through leaf-on (JJA) phenology. Middle row: AR-flow per-quarter CHM predictions. Bottom: per-tile AR-flow MAE  across a 50-tile pilot. This pilot differs from the  789-tile frozen-Anchor evaluation in Table~\ref{tab:seasonal}. }
\label{fig:seasonal}
\end{figure}

\section{Cost Comparison}
\label{sec:cost_appendix}

\begin{table}[H]
\centering
\caption{Cost comparison: LiDAR vs commercial vs ours.}
\fitwidth{
\begin{tabular}{lccccl}
\toprule
Method & \$/km$^2$ & CONUS & Global & Cadence & Access \\
\midrule
Airborne LiDAR (3DEP) & \$100--300 & \$0.8--2.4\,B & \$15--45\,B & multi-year & open \\
Maxar Vivid2 + CHMv2 & commercial & commercial & commercial & annual & gated by license \\
\textbf{Ours (forest-extent)} & \textbf{$<$\$0.01} & \textbf{\$6--9K} & \textbf{\$75--115K} & \textbf{quarterly} & \textbf{open} \\
\bottomrule
\end{tabular}}
\label{tab:cost}
\end{table}

Forest-extent CONUS (\textasciitilde12~M tiles) requires \textasciitilde4,300 A100-hours per pass; global (\textasciitilde160~M tiles at 1.3~s/tile) requires \textasciitilde58,000 A100-hours per pass. Costs assume on-demand mid-market GPU pricing (\$1.30--2.00/A100-hr).

\section{Failure Modes}
\label{sec:failure_appendix}

\begin{figure}[H]
\centering
\includegraphics[width=\linewidth]{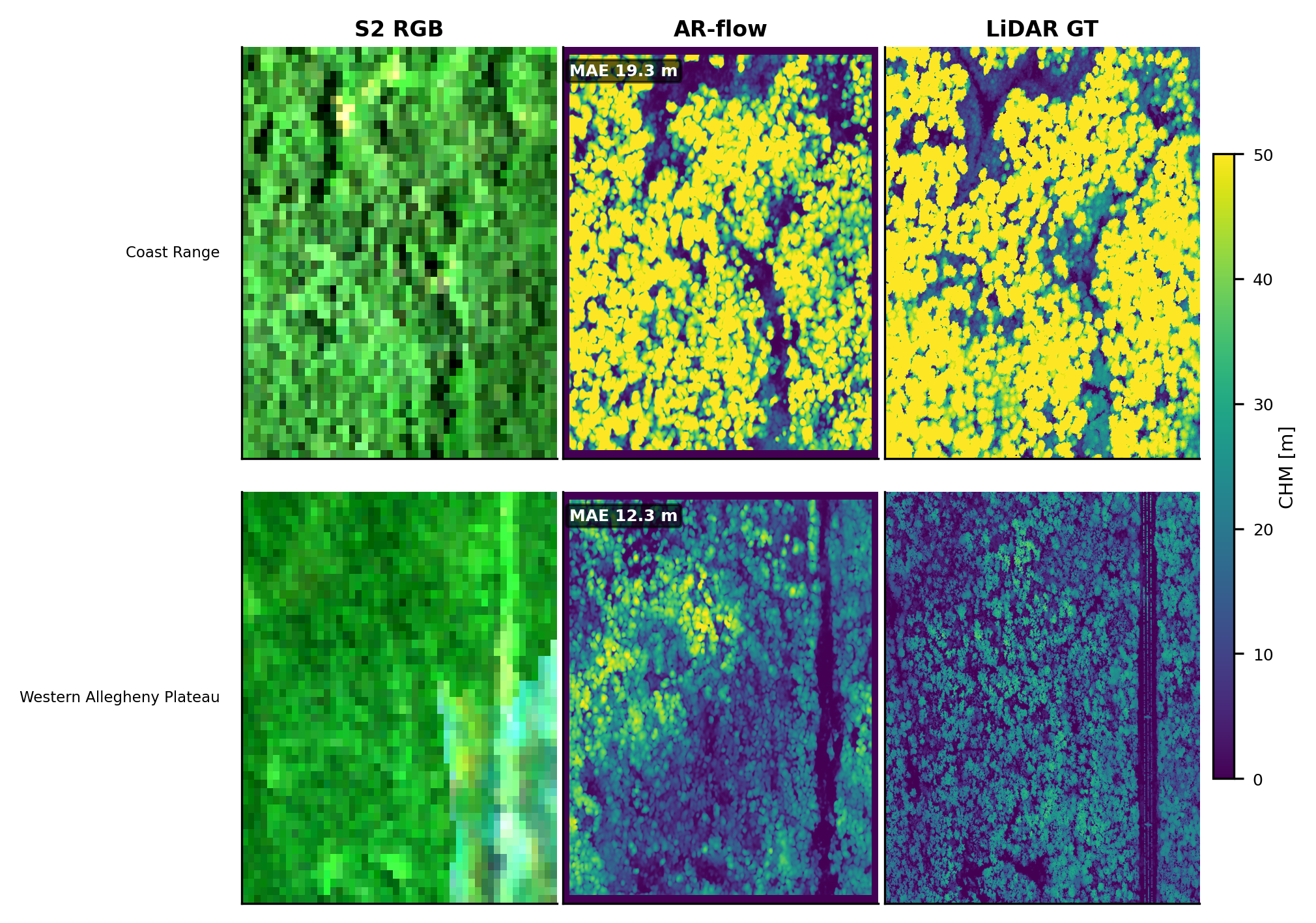}
\caption{\textbf{Failure modes.} Two worst-case AR-flow tiles from distinct biomes (season-matched JJA). \textbf{Top}: Coast Range tall conifer (MAE 19.3~m, ground-truth mean 44~m) -- systematic underestimation of dense old-growth canopy, the dominant failure mode in the cohort. \textbf{Bottom}: Western Allegheny Plateau eastern deciduous (MAE 12.3~m, ground-truth mean 13~m) -- mid-canopy overestimation with visible LiDAR flight-line seam artifacts reproduced in the prediction. Columns: Sentinel-2 RGB, AR-flow prediction, LiDAR ground truth (shared viridis 0--50~m).}
\label{fig:failure_modes}
\end{figure}

\section{Hurricane Helene Exploratory Wind-Disturbance Probe ($n=3$)}
\label{sec:helene}

This is an exploratory probe of the alarm channel under wind disturbance: $n=3$ out-of-cohort tiles with no LiDAR ground truth, using HiForm's S2-NDVI-derived high-severity damage layer as a spatial proxy mask. Hurricane Helene, NC-SC-TN-VA, Sept 26 2024. The same AR-flow checkpoint produced pre (SON 2023) and post (SON 2024) sub-meter CHM rasters.

\begin{table}[h]
\centering
\begin{tabular}{lrrrr}
\toprule
tile\_id & HiForm acres & $\overline{\Delta\text{CHM}}$ damage (m) & $\overline{\Delta\text{CHM}}$ out (m) & damage $-$ out (m) \\
\midrule
6440\_19521 & 59.7 & $-8.13$ & $-3.85$ & $-4.28$ \\
6587\_19647 & 54.6 & $-2.16$ & $-2.99$ & $+0.83$ \\
6535\_19561 & 35.6 & $-1.14$ & $+0.52$ & $-1.67$ \\
\textbf{mean} & \textbf{50.0} & $\mathbf{-3.81}$ & $\mathbf{-2.11}$ & $\mathbf{-1.70}$ \\
\bottomrule
\end{tabular}
\caption{Hurricane Helene exploratory $\Delta$CHM probe.}
\label{tab:helene}
\end{table}

The model's $\Delta$CHM goes anomalously negative inside the HiForm-flagged footprint (mean $-3.81$~m vs $-2.11$~m outside, differential $-1.70$~m). With $n=3$ and no LiDAR truth, this is positioning evidence for the alarm channel under wind disturbance, not a calibrated measurement.

\begin{figure}[H]
\centering
\includegraphics[width=\linewidth]{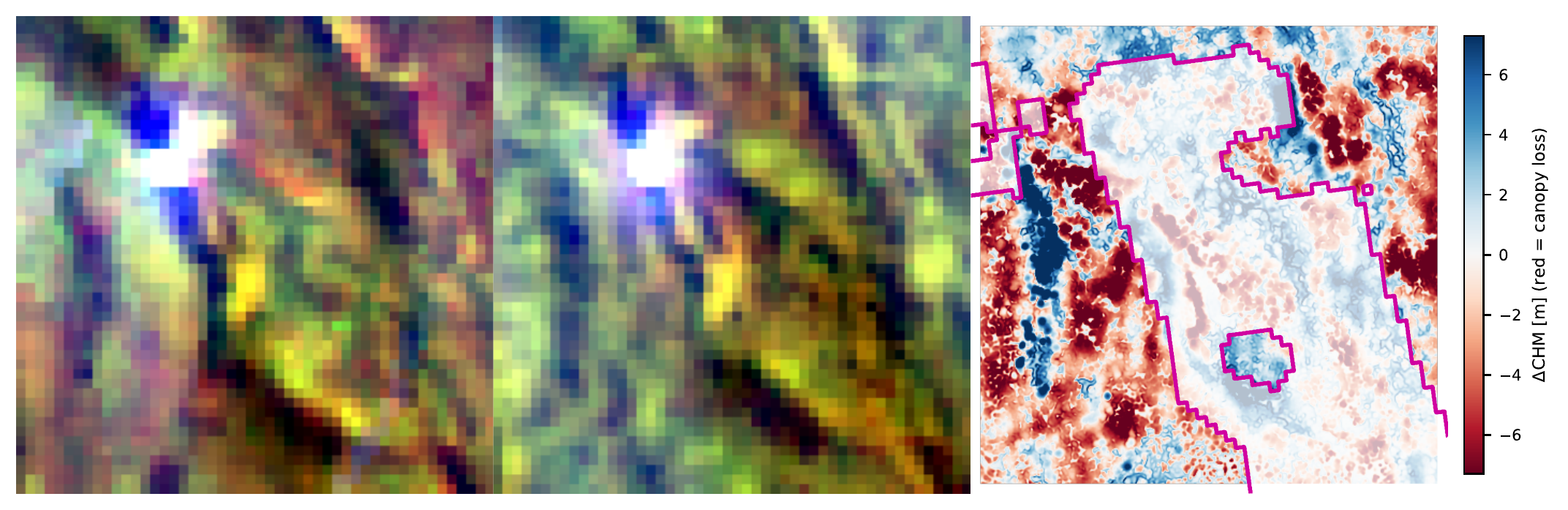}
\caption{ \textbf{Hurricane Helene exploratory change signal} ( tile \texttt{6535\_19561} ).  Left to right:  Sentinel-2 RGB  before Helene (SON 2023) , RGB  after Helene (SON 2024) , and AR-flow $\Delta$CHM  (post minus pre; red  indicates loss ) . The magenta contour marks the HiForm dNDVI-derived damage  proxy;  pixels outside  it are faded.  Mean predicted change  is  $-1.14$~m  inside and $+0.52$~m outside.  Across all three tiles, the means are $-3.81$  and $-2.11$~m.  No pre/post LiDAR is available: these are predicted changes within a proxy damage mask, not  measured height loss or verified tree-fall  locations.}
\label{fig:helene}
\end{figure}

\section{ \texorpdfstring{Cross-LiDAR Validation: NEON Airborne (\(n=8\)) and GEDI Spaceborne (\(n=2{,}526\))}{Cross-LiDAR Validation: NEON Airborne (n=8) and GEDI Spaceborne (n=2\{,\}526)}}

\label{sec:cross_lidar}

We extend the single-truth 3DEP comparison to two independent LiDAR sources: NEON AOP (research-grade \textasciitilde1~m airborne, independent from 3DEP) and GEDI L2A RH98 (spaceborne 25~m footprint, sparse sampling).

\textbf{NEON ( \(n=8\), year-aligned).} 8 tiles from the test cohort falling inside NEON-flown sub-quadrants with exact year alignment, spanning 5 EPA-L3 ecoregions.

\textbf{GEDI ( \(n=2{,}526\)).} GEDI L2A RH98 rasterised onto the 0.5~m grid; 2,526 of 2,542 tiles cleared the 100-pixel threshold.

 \begin{table}[H]
\centering
\caption{ \textbf{Cross-LiDAR validation.} Pearson $r$ is averaged over tiles. Each $\Delta$ compares with 3DEP on the corresponding pixels, not with the full-cohort MAE in Table~\ref{tab:headline}. Rasterizing GEDI onto 0.5~m cells does not create sub-meter reference measurements.}
 \fitwidth{
\begin{tabular}{llrrrrr}
\toprule
Model & Reference truth & $n$ tiles & MAE (m) & RMSE (m) & Pearson $r$ & $\Delta$ MAE vs 3DEP (m) \\
\midrule
Anchor & NEON airborne (\textasciitilde1~m) & 8 & $1.79$ & $2.84$ & $0.437$ & $+0.02$ \\
AR-flow & NEON airborne (\textasciitilde1~m) & 8 & $2.27$ & $3.75$ & $0.279$ & $+0.04$ \\
Anchor & GEDI RH98 spaceborne (25~m) & 2,526 & $4.49$ & $5.74$ & $0.100$ & $+1.67$ \\
AR-flow & GEDI RH98 spaceborne (25~m) & 2,526 & $5.30$ & $7.04$ & $0.067$ & $+1.64$ \\
\bottomrule
\end{tabular}}
\label{tab:cross_lidar}
\end{table}

The NEON  errors are similar to those against 3DEP on the same pixels: MAE increases by 0.02~m for the Anchor and 0.04~m for AR-flow in this eight-tile subset. GEDI uses a different height statistic (RH98 ) and sparse 25~m  footprints. Its errors therefore describe agreement with a different reference and should not be compared directly with the primary 3DEP MAEs.

\subsection{GEDI Agreement by Aggregation Scale}

\label{sec:gedi_scales}

The GEDI correlations in Table~\ref{tab:cross_lidar} average per-tile correlations. Table~\ref{tab:gedi_scales} instead pools valid samples and varies their aggregation scale, so its values are not directly comparable with those tile averages. Agreement rises with aggregation scale for our models and for airborne 3DEP alike. At the 25~m footprint scale the Anchor (0.44) is within 0.02 of 3DEP (0.46), and at 500~m both reach 0.87--0.89. The scale dependence is consistent with a substantial role for footprint and sampling differences; it does not isolate their contribution from other errors.

\begin{table}[H]
\centering
\caption{\textbf{GEDI agreement by aggregation scale} (Pearson $r$, higher better; pooled valid samples, with counts varying by model pair and scale).}
\begin{tabular}{lrrr}
\toprule
Scale & Anchor vs GEDI & AR-flow vs GEDI & GEDI vs 3DEP \\
\midrule
Per-pixel (pooled) & 0.43 & 0.38 & 0.40 \\
25~m footprint disc & 0.44 & 0.43 & 0.46 \\
Coarse 100~m & 0.65 & 0.65 & 0.68 \\
Coarse 500~m & \textbf{0.87} & \textbf{0.87} & 0.89 \\
\bottomrule
\end{tabular}
\label{tab:gedi_scales}
\end{table}

\section{Per-Ecoregion Paired-Bootstrap Forest Plot}
\label{sec:forest_plot}

\begin{table}[H]
\centering
\fitwidth{
\begin{tabular}{lrrcrr}
\toprule
Meta-MAE group & n ecos & n tiles & Anchor winrate (MAE) & Mean $\Delta$ MAE (m) \\
\midrule
Low error (Meta MAE $< 1$~m) & 14 & 766 & 3/14 & $+0.065$ (Meta wins) \\
Medium error ($1$ to $3$~m) & 22 & 682 & 6/22 & $+0.089$ (Meta wins) \\
\textbf{High error (Meta MAE $\geq 3$~m)} & \textbf{33} & \textbf{1,094} & \textbf{18/33} & $\mathbf{-0.258}$ \textbf{(Anchor wins)} \\
\bottomrule
\end{tabular}}
\caption{Per-ecoregion MAE grouped by Meta MAE, not measured canopy height.}
\label{tab:canopy_bin}
\end{table}

\begin{figure}[H]
\centering
\includegraphics[width=\linewidth]{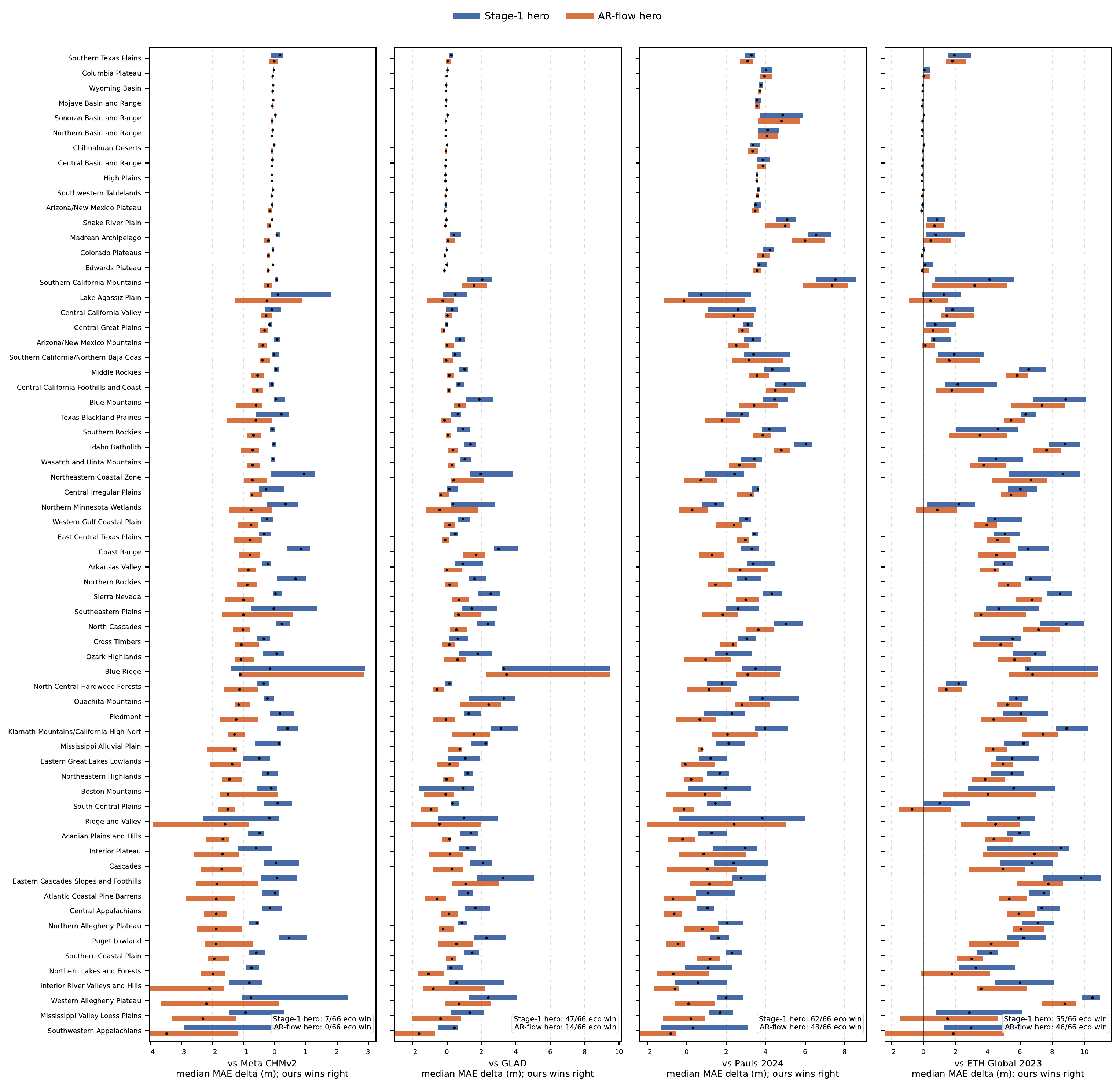}
\caption{\textbf{Per-EPA-L3-ecoregion paired-bootstrap MAE delta vs four competitors.} Each panel one competitor (Meta CHMv2 / GLAD / \citealt{pauls2024canopy} / ETH Global). Per ecoregion, two horizontal 95\% CI bars: Anchor (blue) and AR-flow (orange). Delta = competitor $-$ ours;  positive values favor our model. The inset win counts require the full 95\% interval to lie above zero. This stricter criterion on 66 ecoregions ($n \geq 3$ tiles)  differs from the regional mean-MAE counts in Section~\ref{sec:eco_analysis}. Regions are ordered by AR-flow vs Meta median delta. }
\label{fig:winrate}
\end{figure}

\section{ Spatial Holdout within Known Ecoregions}

\label{sec:ood}

Five EPA Level-III ecoregions were excluded during dataset construction (Central Corn Belt Plains, Driftless Area, Eastern Corn Belt Plains, Erie Drift Plain, Willamette Valley). A separate 9,041-tile spatial-holdout cohort within in-distribution ecoregions (tiles  \(>\)5~km from any train/test tile)  has a reported Anchor MAE of 2.62~m. Its different geographic composition prevents a direct accuracy comparison with the primary cohort. The five truly unseen ecoregions remain untested.

 \begin{table}[H]
\centering
 \fitwidth{
\begin{tabular}{llrrrrl}
\toprule
Ecoregion & L2 biome & $n$ & LiDAR $\bar{h}$ (m) & Anchor MAE (m) & Class \\
\midrule
Coast Range & Marine West Coast Forest & 30 & 13.9 & 6.12 & meaningful \\
Klamath Mtns / CA High N.\ Coast & Western Cordillera & 48 & 15.4 & 6.30 & meaningful \\
Northern Rockies & Western Cordillera & 40 & 10.2 & 5.26 & meaningful \\
Interior Plateau & SE USA Plains & 39 & 9.8 & 5.18 & meaningful \\
Idaho Batholith & Western Cordillera & 46 & 6.2 & 4.55 & meaningful \\
Middle Rockies & Western Cordillera & 39 & 5.4 & 4.18 & meaningful \\
Eastern Cascades Slopes & Western Cordillera & 12 & 7.7 & 3.56 & meaningful \\
Southern Rockies & Western Cordillera & 46 & 4.9 & 3.21 & meaningful \\
Sierra Nevada & Western Cordillera & 43 & 4.5 & 3.00 & meaningful \\
AZ/NM Mountains & Upper Gila Mountains & 47 & 4.3 & 2.96 & meaningful \\
Central CA Foothills & Mediterranean California & 49 & 2.9 & 1.66 & meaningful \\
Colorado Plateaus & Cold Deserts & 49 & 2.0 & 1.61 & near-zero \\
AZ/NM Plateau & Cold Deserts & 48 & 1.4 & 1.33 & near-zero \\
Southern CA Mountains & Mediterranean California & 19 & 2.5 & 1.04 & meaningful \\
Central Basin and Range & Cold Deserts & 47 & 0.4 & 0.42 & near-zero \\
Southwestern Tablelands & S.\ Central Semi-Arid Prairies & 49 & 0.4 & 0.38 & near-zero \\
Wyoming Basin & Cold Deserts & 49 & 0.2 & 0.30 & near-zero \\
High Plains & S.\ Central Semi-Arid Prairies & 47 & 0.3 & 0.28 & near-zero \\
Mojave Basin and Range & Warm Deserts & 42 & 0.1 & 0.12 & near-zero \\
\bottomrule
\end{tabular}}
\caption{ Reported spatial-holdout results by ecoregion. The listed rows total 789 tiles; the aggregate 9,041-tile result is reported separately and cannot be reconstructed from these rows.}
\label{tab:ood_spatial}
\end{table}

\section{Block R$^2$ (\citealt{tolan2024canopy} Protocol)}
\label{sec:block_r2}

Adapted from \citet{tolan2024canopy}, whose protocol Meta CHMv2 also uses, we compute Block R$^2$ on 50$\times$50-pixel non-overlapping blocks and average over tiles, excluding tiles with $<$5\% vegetation cover.

\begin{table}[h]
\centering
\caption{\textbf{Cohort-mean Block R$^2$.} 1,844--1,847 tiles scored (varies by product availability), remaining tiles skipped (low vegetation).}
\begin{tabular}{lrr}
\toprule
Model & Block R$^2$ & $n$ valid \\
\midrule
\textbf{Anchor} & $\mathbf{0.287}$ & $1{,}844$ \\
AR-flow & $-1.198$ & $1{,}844$ \\
Meta CHMv2 & $-0.002$ & $1{,}847$ \\
\bottomrule
\end{tabular}
\label{tab:block_r2}
\end{table}

The Anchor  has the highest mean Block R\(^2\). AR-flow's negative  score shows that realistic fine-scale variation does not ensure accurate block-level heights. The latent-space MSE objective does not itself guarantee optimal decoded Block R\(^2\).

\section{Temporal Fairness: Vintage-Matched Comparison}

\label{sec:vintage}

 Competitor products use imagery from different  periods. We compare the full cohort with a LiDAR 2019--2020 subset  in one archived re-scoring run, using the same scorer for both cohorts. The subset reduces some acquisition-date differences but does not exactly match every product's vintage (Table~\ref{tab:vintages}). Counts vary with valid product coverage.

 \begin{table}[H]
\centering
\caption{\textbf{Competitor imagery vintages.}}
 \begin{tabular}{llp{5cm}}
\toprule
Product & Input imagery & Nominal vintage \\
\midrule
Our pipeline (Anchor / AR-flow) & Sentinel-2 L2A & year-aligned (every tile) \\
Meta CHMv2 \citep{brandt2026meta} & Maxar Vivid2 & 2017--2020 (\textasciitilde80\% 2018--2020) \\
ETH-S2 \citep{lang2023high} & Sentinel-2 (multi-date, fused) & \textasciitilde2020 \\
GLAD \citep{potapov2021glad} & Landsat & \textasciitilde2019 \\
LandFire FCH \citep{landfire2023} & Landsat, lidar and plot-trained models & LF 2023 update (disturbance through 2023) \\
GEDI RH98 \citep{dubayah2020gedi} & GEDI (\textasciitilde25~m footprints) & 2019 onward (sparse) \\
Simard \citep{simard2011canopy} & ICESat/GLAS + ancillary maps & 2005 \\
\citet{pauls2024canopy} & S1 + S2 & 2020 \\
EchoSat \citep{pauls2026echosat} & S1 + S2 + PALSAR-2 + TanDEM-X DEM & annual, 2018--2024 \\
\bottomrule
\end{tabular}
\label{tab:vintages}
\end{table}

 \begin{table}[H]
\centering
\caption{ \textbf{LiDAR-vintage sensitivity within one re-scoring run.} All heights are evaluated on the common 0.5~m grid. MAEs are tile averages; counts vary by product coverage. These independently re-scored values should be compared within this table,  not substituted for the headline results.}
 \fitwidth{
\begin{tabular}{lrrrrr}
\toprule
& \multicolumn{2}{c}{Full cohort} & \multicolumn{3}{c}{LiDAR 2019--2020} \\
\cmidrule(lr){2-3}\cmidrule(lr){4-6}
Product & $n$ & MAE (m) & $n$ & MAE (m) & RMSE (m) \\
\midrule
Anchor & 2,530 & 2.820 & 911 & 2.680 & 3.800 \\
AR-flow & 2,530 & 3.664 & 911 & 3.486 & 5.105 \\
Meta CHMv2 & 2,530 & 2.896 & 911 & 2.898 & 4.259 \\
GLAD & 2,527 & 4.083 & 908 & 3.974 & 5.184 \\
LandFire FCH & 2,530 & 4.709 & 911 & 4.832 & 6.277 \\
ETH Global & 2,527 & 7.048 & 908 & 7.003 & 8.334 \\
EchoSat & 2,530 & 6.221 & 911 & 6.448 & 7.545 \\
Pauls / AI4Forest & 2,530 & 6.649 & 911 & 6.885 & 7.710 \\
\bottomrule
\end{tabular}}
\label{tab:vintage_matched}
\end{table}

 Within this re-scoring run, the Anchor retains lower MAE than Meta on the 2019--2020 subset (2.680 versus 2.898~m). AR-flow  reaches 3.486~m. These results concern height error; the saved re-scoring record does not include perceptual metrics.

\section{ Exploratory Wildfire  Change Signal}

\label{sec:disturbance}

The CARR cohort ( \(n=6\) high-severity-overlap tiles) is illustrative single-event evidence supporting the alarm-channel framing in Section~\ref{sec:discussion}.

\textbf{CARR Fire (Shasta County, CA, 2018).}  AR-flow  predicts strongly negative in-burn  \(\Delta\)CHM on four of six high-severity-overlap tiles. On tile \texttt{9042\_12258}, the mean predicted in-burn  change is \(-17.1\)~m. This is a model-derived difference, not a measured height-loss estimate. The example has post-fire LiDAR but no pre-fire LiDAR, so it cannot establish the fraction of true height loss recovered.

\begin{figure}[H]
\centering
\includegraphics[width=\linewidth]{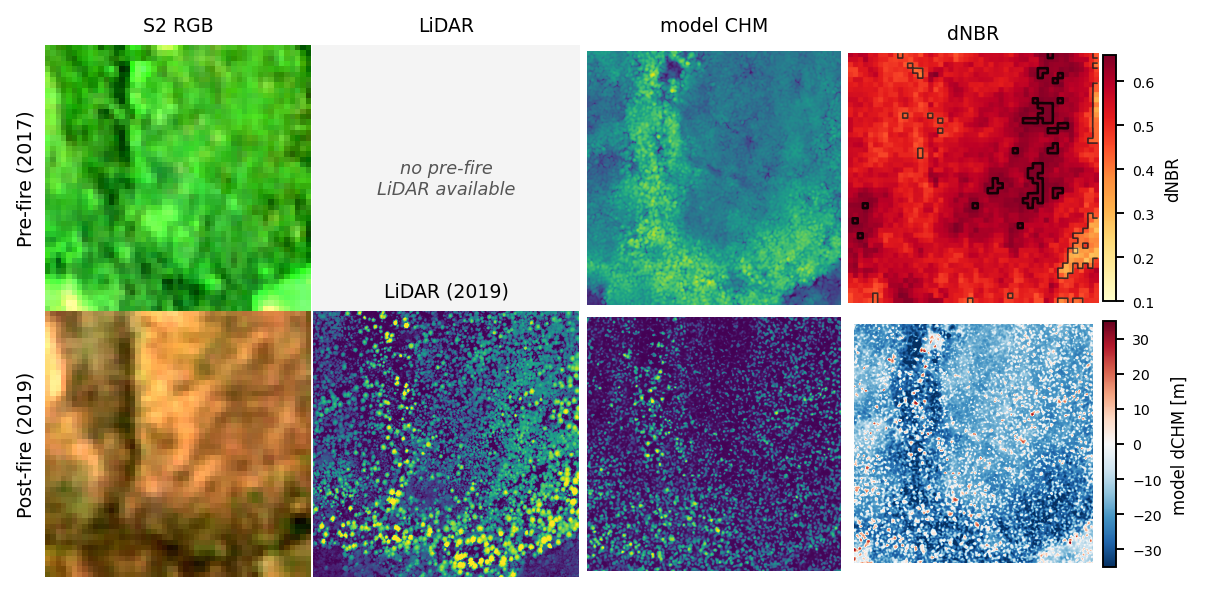}
\caption{ \textbf{Carr Fire 2018 change example} (tile \texttt{9042\_12258}, Shasta County, CA ).  Top: pre-fire Sentinel-2 RGB  and AR-flow  CHM  ( 2017),  a placeholder for unavailable pre-fire LiDAR,  and dNBR fire severity (\citealt{key2006nbr} ).  Bottom:  post-fire RGB, LiDAR, and AR-flow CHM (2019 ) , followed by predicted $\Delta$CHM  (post  minus pre; blue indicates loss, red  gain).  Post-fire LiDAR  supports a height comparison at one date; without a pre-fire reference,  the  figure does not validate the magnitude of height change.}
\label{fig:disturbance}
\end{figure}

The  predicted decrease illustrates a possible disturbance-screening signal. It does not quantify biomass loss; calibrated  change estimation requires a cohort with paired pre/post reference measurements.

\section{Intermediate Operating Points}

\begin{figure}[H]
\centering
\includegraphics[width=0.8\linewidth]{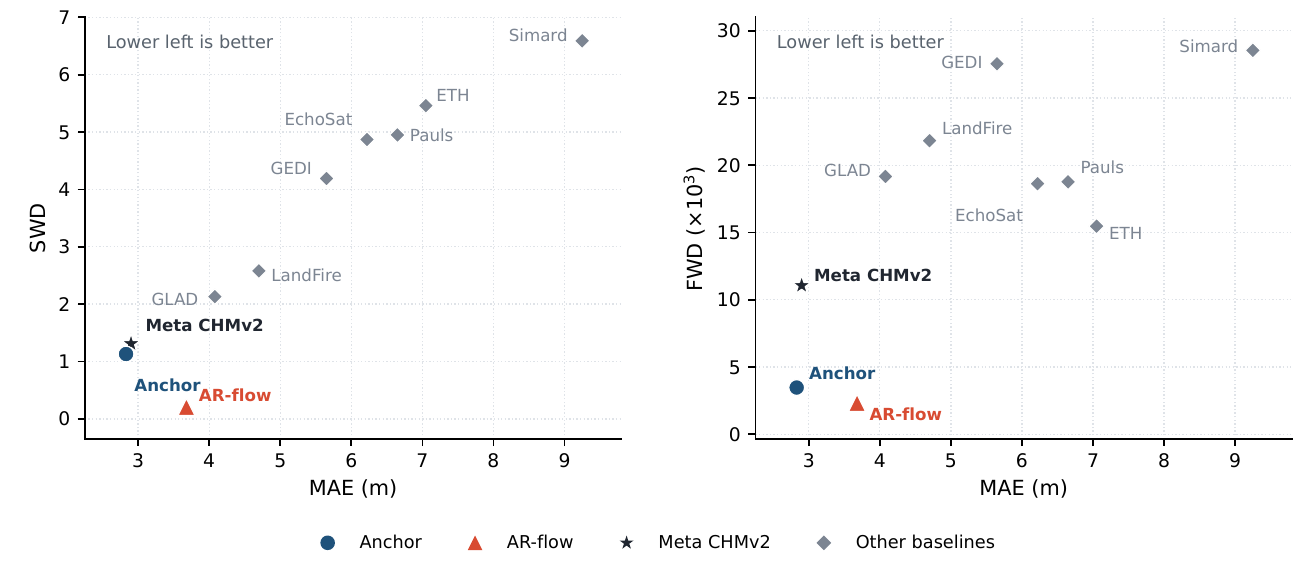}
\caption{ \textbf{Distortion-perception comparison from Table~\ref{tab:sota}(a).} Left panel y = SWD; right panel y = FWD ($\times 10^3$); shared x-axis = MAE (m). Anchor (navy circle), AR-flow (red triangle), Meta CHMv2 (black star), free-input baselines (gray diamonds). Lower-left is better.}
\label{fig:pareto}
\end{figure}
\label{sec:intermediate_appendix}

Linear interpolation between Anchor and AR-flow outputs in pixel space, $\hat{y}_\alpha = (1-\alpha)\cdot\text{Anchor} + \alpha\cdot\text{AR-flow}$, provides intermediate operating points between the two outputs. Table~\ref{tab:intermediate} reports cohort-level metrics for six values of $\alpha$ on all 2,530 test tiles.

\begin{table}[H]
\centering
\caption{Intermediate operating points via Anchor/AR-flow interpolation ($n{=}2{,}530$ tiles). EE uses raw Sobel magnitude L1 (unnormalized); absolute values differ from Table~\ref{tab:headline} which uses per-tile-max normalization, and their endpoint ordering differs: raw edge L1 decreases while normalized EE increases.}
\begin{tabular}{lrrr}
\toprule
$\alpha$ & MAE (m) & RMSE (m) & EE \\
\midrule
0.0 (pure Anchor) & \textbf{2.83} & \textbf{4.06} & 0.557 \\
0.2 & 2.86 & 4.11 & 0.458 \\
0.4 & 2.98 & 4.29 & 0.381 \\
0.6 & 3.15 & 4.59 & 0.334 \\
0.8 & 3.39 & 4.98 & 0.306 \\
1.0 (pure AR-flow) & 3.68 & 5.43 & \textbf{0.289} \\
\bottomrule
\end{tabular}
\label{tab:intermediate}
\end{table}

Across the reported blend weights, MAE increases from 2.83 to 3.68~m while edge error decreases from 0.557 to 0.289 as the interpolation weight shifts from distortion-optimal (Anchor) to perception-optimal (AR-flow). Practitioners can select any point on this curve by blending the two outputs at inference time with no additional training.

\section{Preliminary Uncertainty Quantification}
\label{sec:uq_appendix}

Because the AR-flow is stochastic (starting from random noise), drawing $N$ independent samples with the frozen Anchor yields per-pixel uncertainty estimates at the cost of $N$ Stage-2 forward passes only (the Anchor is computed once). We draw $N{=}8$ samples on 50 randomly selected test tiles and compute per-pixel standard deviation across samples.

\begin{table}[H]
\centering
\caption{Uncertainty quantification from 8 AR-flow samples ($n{=}50$ tiles).}
\begin{tabular}{lr}
\toprule
Statistic & Value \\
\midrule
Mean per-pixel std & 2.28~m \\
Pearson $r$($\sigma$, $|$error$|$) & 0.391 \\
Spearman $\rho$($\sigma$, $|$error$|$) & 0.414 \\
Empirical coverage ($|$error$| \leq 2\sigma$) & 65.7\% \\
\bottomrule
\end{tabular}
\label{tab:uq}
\end{table}

Per-pixel uncertainty correlates positively with actual absolute error (Spearman $\rho{=}0.41$), confirming that multi-sample AR-flow draws provide a meaningful uncertainty signal. The model is moderately overconfident: 65.7\% of pixels fall within $2\sigma$ vs the 95\% expected under Gaussian calibration. Tall-canopy ecoregions show higher per-pixel std (consistent with higher prediction difficulty), while arid/low-canopy ecoregions show near-zero std. Calibrated prediction intervals via conformal methods or temperature scaling are a natural extension.

\section{Zero-Shot Out-of-Distribution Probe: Norway, British Columbia and New Zealand}
\label{sec:ood_probe}

To test whether the CONUS-trained system transfers to unseen biomes, we run zero-shot inference on 12 tiles from three non-CONUS regions: 4 boreal tiles from Norway, 4 temperate/boreal tiles from British Columbia (Canada), and 4 temperate rainforest tiles from New Zealand. LiDAR CHM ground truth is derived from national open-data programs (Norway: hoydedata.no at 1~m; BC: LidarBC at 1~m; NZ: LINZ at 1~m). Sentinel-2 JJA 2023 composites are generated using the same STAC-based pipeline as the training data. No fine-tuning or adaptation is performed.

\begin{table}[H]
\centering
\caption{Zero-shot OOD inference on non-CONUS tiles (AR-flow MAE in meters).}
\begin{tabular}{llrr}
\toprule
Tile & Region / forest type & GT mean (m) & AR-flow MAE (m) \\
\midrule
Saltdal & Norway, boreal pine/birch & 4.0 & 4.29 \\
Steinkjer & Norway, boreal spruce & 3.9 & 4.03 \\
Ringerike & Norway, mixed boreal & 4.1 & 7.74 \\
Elverum & Norway, boreal spruce/pine & 4.9 & 5.67 \\
\midrule
092g003 & BC, coastal Douglas-fir & 4.6 & 2.12 \\
082f008 & BC, interior cedar-hemlock & 7.3 & 6.80 \\
093b065 & BC, sub-boreal spruce & 1.4 & 2.18 \\
094a069 & BC, boreal white/black spruce & 5.1 & 3.38 \\
\midrule
West Coast & NZ, podocarp/broadleaf rainforest & 16.9 & 12.47 \\
Bay of Plenty & NZ, planted radiata pine & 24.9 & 12.75 \\
Kaimai Range & NZ, native bush & 11.2 & 7.80 \\
Fiordland & NZ, beech forest & 17.7 & 14.49 \\
\midrule
\multicolumn{2}{l}{\textbf{Norway mean ($n{=}4$)}} & & \textbf{5.43} \\
\multicolumn{2}{l}{\textbf{BC mean ($n{=}4$)}} & & \textbf{3.62} \\
\multicolumn{2}{l}{\textbf{NZ mean ($n{=}4$)}} & & 11.88 \\
\bottomrule
\end{tabular}
\label{tab:ood}
\end{table}

 Mean MAE is 3.62~m  in BC, 5.43~m  in Norway, and 11.88~m in New Zealand. Each mean covers only four tiles, using separate national LiDAR references at 1~m; these exploratory results do not establish performance equivalent to the CONUS benchmark. New Zealand's tall native rainforest (GT mean 11--25~m, canopy heights up to 60~m) is the hardest OOD regime, with MAE 7.8--14.5~m reflecting the large distributional shift from CONUS training forests. The model captures broad spatial structure in all three regions (clearings, roads, forest edges are correctly predicted as low-canopy) but systematically underpredicts NZ's exceptionally tall canopy. Geographic expansion of the training corpus is the primary path to improving non-CONUS performance.

\section{EPA Level-III Ecoregion Index}
\label{sec:eco_index}

\begin{table}[H]
\centering\small
\caption{\textbf{EPA Level-III ecoregion index.} Map numerals in Figure~\ref{fig:coverage}b. $n$ counts held-out \emph{test} tiles. Of the 85 CONUS ecoregions, 72 appear in the dataset and 13 in neither split; those 13 are drawn unfilled in the figure. A further three carry training tiles but no test tiles, so 16 rows show $n{=}0$ while only 13 are unfilled.}
\label{tab:eco_index}
\resizebox{\linewidth}{!}{%
\begin{tabular}{rlr@{\hskip 2em}rlr}
\toprule
code & ecoregion & $n$ & code & ecoregion & $n$ \\
\midrule
1 & Coast Range & 79 & 44 & Nebraska Sand Hills & 0 \\
2 & Puget Lowland & 31 & 45 & Piedmont & 24 \\
3 & Willamette Valley & 0 & 46 & Northern Glaciated Plains & 0 \\
4 & Cascades & 35 & 47 & Western Corn Belt Plains & 0 \\
5 & Sierra Nevada & 96 & 48 & Lake Agassiz Plain & 6 \\
6 & Central California Foothills and Coastal Mountains & 77 & 49 & Northern Minnesota Wetlands & 6 \\
7 & Central California Valley & 10 & 50 & Northern Lakes and Forests & 18 \\
8 & Southern California Mountains & 24 & 51 & North Central Hardwood Forests & 25 \\
9 & Eastern Cascades Slopes and Foothills & 18 & 52 & Driftless Area & 0 \\
10 & Columbia Plateau & 78 & 53 & Southeastern Wisconsin Till Plains & 0 \\
11 & Blue Mountains & 39 & 54 & Central Corn Belt Plains & 0 \\
12 & Snake River Plain & 50 & 55 & Eastern Corn Belt Plains & 0 \\
13 & Central Basin and Range & 90 & 56 & Southern Michigan/Northern Indiana Drift Plains & 0 \\
14 & Mojave Basin and Range & 64 & 57 & Huron/Erie Lake Plains & 0 \\
15 & Northern Rockies & 88 & 58 & Northeastern Highlands & 44 \\
16 & Idaho Batholith & 90 & 59 & Northeastern Coastal Zone & 10 \\
17 & Middle Rockies & 99 & 60 & Northern Allegheny Plateau & 15 \\
18 & Wyoming Basin & 100 & 61 & Erie Drift Plain & 0 \\
19 & Wasatch and Uinta Mountains & 96 & 62 & North Central Appalachians & 2 \\
20 & Colorado Plateaus & 73 & 63 & Middle Atlantic Coastal Plain & 1 \\
21 & Southern Rockies & 55 & 64 & Northern Piedmont & 0 \\
22 & Arizona/New Mexico Plateau & 45 & 65 & Southeastern Plains & 19 \\
23 & Arizona/New Mexico Mountains & 54 & 66 & Blue Ridge & 3 \\
24 & Chihuahuan Deserts & 29 & 67 & Ridge and Valley & 3 \\
25 & High Plains & 100 & 68 & Southwestern Appalachians & 3 \\
26 & Southwestern Tablelands & 79 & 69 & Central Appalachians & 36 \\
27 & Central Great Plains & 25 & 70 & Western Allegheny Plateau & 7 \\
28 & Flint Hills & 0 & 71 & Interior Plateau & 16 \\
29 & Cross Timbers & 25 & 72 & Interior River Valleys and Hills & 9 \\
30 & Edwards Plateau & 38 & 73 & Mississippi Alluvial Plain & 3 \\
31 & Southern Texas Plains & 10 & 74 & Mississippi Valley Loess Plains & 8 \\
32 & Texas Blackland Prairies & 15 & 75 & Southern Coastal Plain & 41 \\
33 & East Central Texas Plains & 15 & 76 & Southern Florida Coastal Plain & 0 \\
34 & Western Gulf Coastal Plain & 15 & 77 & North Cascades & 100 \\
35 & South Central Plains & 47 & 78 & Klamath Mountains/California High North Coast Range & 80 \\
36 & Ouachita Mountains & 15 & 79 & Madrean Archipelago & 35 \\
37 & Arkansas Valley & 25 & 80 & Northern Basin and Range & 34 \\
38 & Boston Mountains & 8 & 81 & Sonoran Basin and Range & 24 \\
39 & Ozark Highlands & 21 & 82 & Acadian Plains and Hills & 27 \\
40 & Central Irregular Plains & 15 & 83 & Eastern Great Lakes Lowlands & 13 \\
41 & Canadian Rockies & 0 & 84 & Atlantic Coastal Pine Barrens & 15 \\
42 & Northwestern Glaciated Plains & 0 & 85 & Southern California/Northern Baja Coast & 41 \\
43 & Northwestern Great Plains & 1 & & & \\
\bottomrule
\end{tabular}}
\end{table}

\section{Data Availability}\label{sec:data_availability}
Predictions, LiDAR references, and per-tile metrics are intended for release under CC-BY-4.0 through a DOI-bearing archive. Model weights remain withheld because of institutional IP.

\end{document}